%% file: main.tex
\documentclass{article}

\usepackage[preprint]{corl_2026} % Uncomment for pre-prints (e.g., arxiv); This is like ``final'', but will remove the CORL footnote.

\usepackage{booktabs}
\usepackage{makecell}
\usepackage{amsmath}
\usepackage{amssymb} % for \checkmark
\usepackage{pifont}  % optional for \xmark
\usepackage{graphicx}
\usepackage{wrapfig}
\usepackage{makecell}

\newcommand{\method}{COWBOY}

\usepackage[all]{hypcap}

\usepackage[table]{xcolor}
\usepackage{tabularx}
\usepackage{array}
\usepackage{makecell}

\newcolumntype{Y}{>{\centering\arraybackslash}X}

\definecolor{perfGood}{RGB}{239,250,242}
\definecolor{perfMid}{RGB}{250,248,241}
\definecolor{perfBad}{RGB}{250,242,242}
\newcommand{\good}[1]{\cellcolor{perfGood}\makebox[\linewidth][c]{#1}}
\newcommand{\midperf}[1]{\cellcolor{perfMid}\makebox[\linewidth][c]{#1}}
\newcommand{\bad}[1]{\cellcolor{perfBad}\makebox[\linewidth][c]{#1}}

\title{\method: A Scalable Sim-to-Real Framework for Learning Contextual Whole-Body Manipulation}

\author{\textbf{Jiahui Yang$^{\ast}$\qquad Jason Jingzhou Liu$^{\ast}$\qquad Ray Muxin Liu$^{\ast}$} \\
\textbf{Andrew E. Wang\qquad Kenneth Shaw\qquad}
\textbf{Deepak Pathak$^{\dagger}$\qquad Ruslan Salakhutdinov$^{\dagger}$}\\
[0.3em]
Carnegie Mellon University\\ 
$^{\ast}$Equal contribution\qquad $^{\dagger}$Equal advising}

\begin{document}
\maketitle

%===============================================================================

\vspace{-10mm}
\begin{abstract}
Mobile manipulators must adapt not only to objects, but also to the spatial context in which they appear: where to position the base, how to manipulate, and where to look while navigating surrounding scene geometry. To address this, we present COWBOY, a scalable sim-to-real framework for learning such contextual whole-body manipulation. On whole-body dexterous grasping, COWBOY transfers zero-shot to hundreds of objects and diverse in-the-wild scenes, spanning cluttered tabletops, shelves, drawers, and other unstructured settings. 
Rather than learning scene adaptation directly with RL in simulation, a vectorized environment-aware whole-body controller (WBC) handles reaching in thousands of scenes by coordinating base-arm motion, collision avoidance, and active vision.  Then, local RL experts are trained in local scenes for object interaction and can be reused across thousands of full-scene layouts. We distill these composed WBC--RL teachers into a single vision policy that controls the mobile base, arm, dexterous hand, and actuated camera from point-cloud observations. The resulting policy exhibits contextual whole-body behavior, automatically adapting base placement, camera viewpoint, arm trajectory, and grasp strategy to observed scene and object geometry. Videos are available at \url{cowboy-manip.com}
\end{abstract}


\begin{figure}[htbp]
    \centering
    \includegraphics[width=1\linewidth]{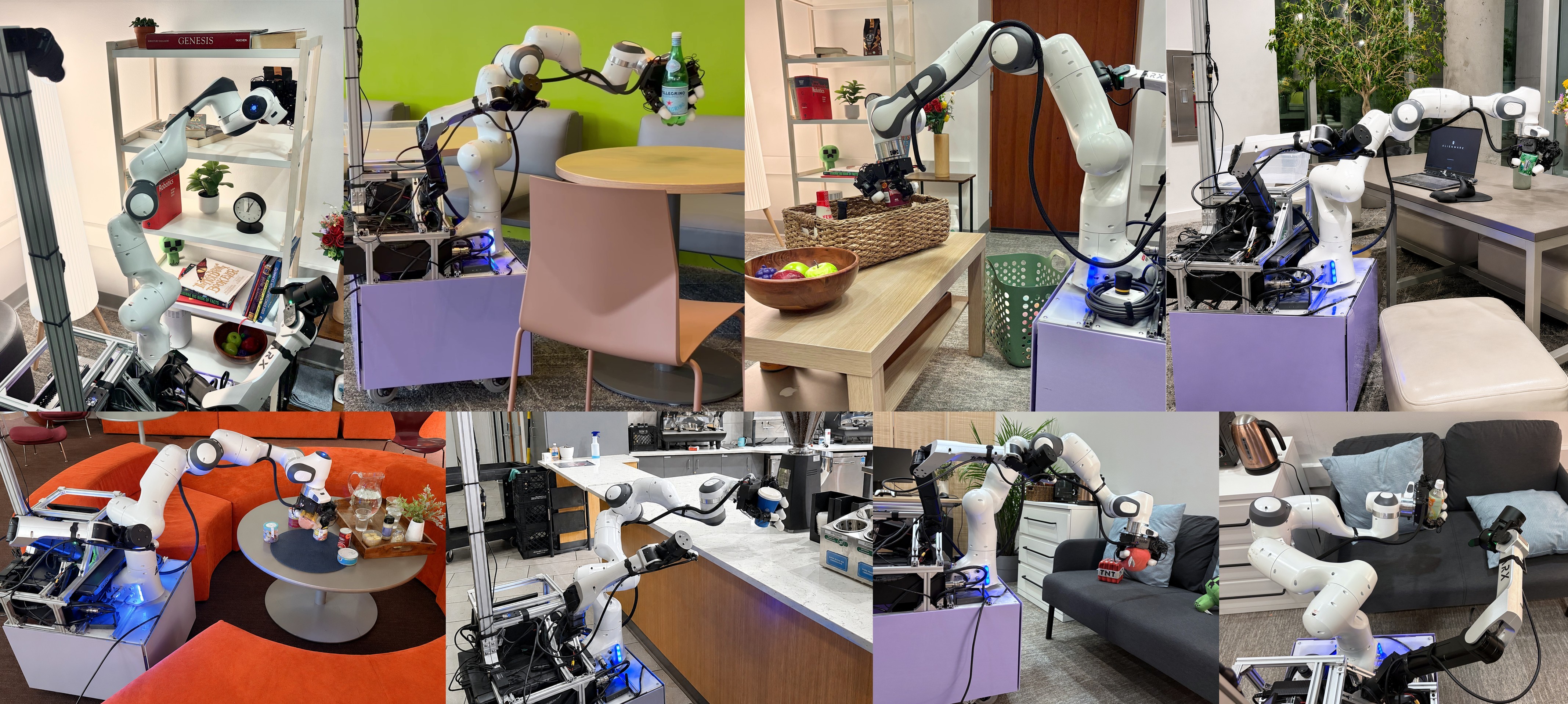}
    \caption{ \small
    COWBOY enables learning whole-body manipulation in simulation across thousands of scenes and objects. We train a single vision-based policy that transfers zero-shot in diverse real-world scenarios across hundreds of objects. The policy adapts its grasping modes, base positioning, camera viewing directions solely depending on the scene context without explicit conditioning.
    }
    \label{fig:teaser}
\end{figure}

%===============================================================================
\enlargethispage{2\baselineskip}

% Talk about generlaization first. Anything in manip are trained and tested in the same env and object set. How does one generalize? Two camps: one camp is training a big model end to end on lots of data in a multi task setup and hoping it generalizes. Issue is as multi task setup, compelxity and data requirement both increase. This kind of setup has yet to show strong signs of genealrizaiton, and has primairly limited to table top scenaiors where the scene is inherently limited by the robot placement.

% In a mobile manip setup, generalization is even harder because you can move to different scenes. Most common approach here is hiearhcyial, but each compoennt of the hierchy does not know of each other so you lose the flexitbility that an end-to-end approach provides. 

% We believe that hierarchy is a strong way to yield genearlization due to the combinatorial nature. Our main insight is that this hierarhcial signal should bee used at trianing time but should be end to end at test time. To achieve this, we propose a framework as the cobination of RL and supervised distillation. 

\vspace{-4mm}
\section{Introduction}
\vspace{-3mm}
Humans effortlessly interact with diverse environments and objects in their daily lives. A person reaching for the same object may move very differently depending on whether it lies on an open tabletop, inside a shelf, behind clutter, or within a drawer. These behaviors rely on whole-body adaptation: people reposition their body, adjust their viewpoint, and coordinate their arms and hands to maintain visibility and avoid obstacles.

Similarly, deploying robots in everyday environments requires policies that generalize not only across objects, but also across the spatial contexts in which those objects appear. For mobile manipulators, this is further complicated by partial observability, since onboard sensors provide only an incomplete view of the scene. From this context alone, the robot must coordinate its base, arm, and camera to reach the object, maintain visibility, and avoid collisions. We refer to this scene-conditioned whole-body reasoning capability as \emph{contextual whole-body manipulation}.

% Learning contextual whole-body manipulation requires training policies across diverse scenarios. Simulation-based RL provides a promising path, since diverse objects and scenes can be generated at scale. Recent sim-to-real manipulation systems have demonstrated impressive generalization across large object sets \cite{dextrahrgb}; however, they typically do not generalize across diverse environments, and are often trained within a single fixed setting \cite{dextrahg}. Training mobile manipulation policies across varied environments remains challenging as it requires global scene understanding that goes beyond local object interactions, often involving high-dimensional spatial representations. This complicates both policy conditioning and reward specification. The difficulty is further exacerbated by the high-dimensional action spaces of mobile manipulators, requiring coordinated control over the base, arm, and perception.

Reinforcement learning (RL) in simulation is a promising approach for learning 
contextual whole-body manipulation as diverse training scenarios can be generated at scale. Recent sim-to-real manipulation systems have demonstrated impressive generalization across large object sets; however, they typically do not generalize across diverse environments and are often trained in a single fixed setting \cite{dextrahg, dextrahrgb}. Training mobile manipulation policies across varied environments remains challenging because it requires global scene understanding beyond local object interactions, often over high-dimensional spatial representations. This complicates both policy conditioning and reward specification. The difficulty is exacerbated by the high-dimensional action spaces of mobile manipulators, requiring coordination of the base, arm, and perception.

To this end, we present \method, a sim-to-real framework for training mobile manipulation skills across thousands of diverse environments and hundreds of objects, enabling contextual whole-body manipulation learning at scale. Our framework first proposes a vectorized, environment-aware whole-body controller (WBC) that operates across many scenes on a single GPU. Given only an end-effector goal, the WBC coordinates base-arm motion to reach the target while avoiding obstacles in the scene. It also controls an actuated neck to maintain visibility to the object.

Our WBC enables us to decompose contextual whole-body manipulation into two stages: global free-space reaching and local interaction. First, the WBC brings the end-effector near a target object while avoiding obstacles and maintaining gaze on the object. Then, for each desired manipulation behavior, we train a local RL policy 
% with WBC task-space commands 
to learn interactions across hundreds of objects. Since each RL policy only needs to handle one type of local manipulation behavior, policy conditioning and reward specification become tractable. Finally, we distill both the WBC reaching behavior and the state-based local policies into a unified vision-based whole-body policy. Importantly, during distillation, we reuse each local policy by generating thousands of full-scene layouts around the local scene, enabling scalable training across diverse scenes and objects.

We demonstrate COWBOY on whole-body dexterous grasping with a mobile manipulator equipped with a 16-DoF hand and an actuated neck for active vision. COWBOY achieves zero-shot sim-to-real transfer, robustly grasping hundreds of geometrically diverse objects across in-the-wild settings, including shelves, drawers, and cluttered tabletops. The policy generates adaptive, context-dependent behaviors, coordinating approach direction, base placement, camera viewpoint, grasp strategy, and collision-avoiding motion from the surrounding scene. Ablations validate the key design choices of our framework and show that they are critical for robust in-the-wild mobile manipulation.

%===============================================================================
\enlargethispage{1\baselineskip}
\vspace{-3mm}
\section{Related Works}
\vspace{-2mm}
% \subsection{Robot Learning for Mobile Manipulation}
% Mobile manipulation has largely advanced through human demonstration-driven pipelines. These methods learn coordinated base, arm, and end-effector behavior from teleoperated or robot-free human demonstrations \cite{mobilealoha, haoyudoor, xiong2025visionactionlearningactive, xu2026hommi, jiang2025behaviorrobotsuitestreamlining}. This paradigm has enabled impressive long-horizon whole-body behaviors, but its scalability depends on collecting demonstrations with sufficient coverage over scenes, tasks, and object instances, often through specialized pipelines such as embodiment-specific teleoperation interfaces \cite{mobilealoha}, VR teleoperation \cite{xiong2025visionactionlearningactive}, or cross-embodiment retargeting from robot-free human demonstrations \cite{xu2026hommi}.
% A complementary direction learns whole-body mobile manipulation in simulation, with sim-to-real results spanning wheeled platforms \cite{spin}, legged robots \cite{liu2024visualwholebodycontrollegged, pmlr-v205-fu23a}, and humanoids \cite{he2025viralvisualsimtorealscale, xue2025openingsimtorealdoorhumanoid}. This line of work establishes the feasibility of simulation-trained whole-body mobile manipulation, but most evaluate generalization in more constrained manipulation settings than the one we consider here. In contrast, we study generalizable mobile dexterous manipulation in heavy clutter, where the robot must coordinate the mobile base, arm, dexterous hand, and active camera arm to grasp a wide range of objects across diverse scenes.
\subsection{Robot Learning for Mobile Manipulation and Whole-Body Control}
\vspace{-3mm}
% Mobile manipulation has largely advanced through demonstration-driven pipelines that learn coordinated base, arm, and end-effector behavior from real-world data. Mobile ALOHA introduced low-cost whole-body teleoperation for bimanual mobile tasks~\cite{mobilealoha}. Adaptive Mobile Manipulation studied open-world articulated-object interaction~\cite{haoyudoor}. Vision in Action learned active perception from human demonstrations~\cite{xiong2025visionactionlearningactive}. Hommi learned whole-body mobile manipulation from human demonstrations~\cite{xu2026hommi}. HoMeR combined imitation learning with whole-body control for in-the-wild household tasks~\cite{sundaresan2026homer}. Behavior Robot Suite streamlined real-world data collection for household whole-body manipulation~\cite{jiang2025behaviorrobotsuitestreamlining}. These systems achieve impressive long-horizon behaviors, but scaling demonstrations is difficult because each trajectory must encode scene-dependent choices such as base placement, approach direction, visibility, and obstacle avoidance. 
Mobile manipulation has largely advanced through demonstration-driven pipelines that learn coordinated base, arm, and end-effector behavior from real-world data, with advances in low-cost whole-body teleoperation~\cite{mobilealoha}, open-world articulated-object interaction~\cite{haoyudoor}, active perception from human demonstrations~\cite{xiong2025visionactionlearningactive}, whole-body imitation from human demonstrations~\cite{xu2026hommi,sundaresan2026homer}, and scalable household data collection~\cite{jiang2025behaviorrobotsuitestreamlining}. Although these systems achieve impressive long-horizon behavior, scaling demonstrations is difficult because each trajectory must encode scene-dependent base placement, approach direction, visibility, and obstacle avoidance.

RL end-to-end whole-body control has become popular for training fast WBC. Beyond direct visual imitation or keypoint tracking frameworks~\cite{Allshireetal2025, Yinetal2025, Xieetal2025}, alternative paradigms leverage residual learning for task-specific manipulation~\cite{Zhaoetal2025}, adversarial motion priors for emergent mode-switching~\cite{LErarioetal2023}, or teleoperation interfaces with real-time coordination~\cite{Heetal2024}.  Recent humanoid systems have further shown visual sim-to-real transfer for loco-manipulation and door opening~\cite{he2025viralvisualsimtorealscale,xue2025openingsimtorealdoorhumanoid}.  Spin studied wheeled mobile manipulation with perception, interaction, and navigation~\cite{spin}. Deep Whole-Body Control learned unified locomotion-manipulation policies for legged robots~\cite{pmlr-v205-fu23a}. Visual Whole-Body Control extended this direction to vision-based legged loco-manipulation~\cite{liu2024visualwholebodycontrollegged}. These works demonstrate the promise of simulation-trained whole-body control, but evaluate generalization in settings with more constrained object--environment variation than ours. N$^2$M$^2$~\cite{honerkamp2023n2m2} and Boosted Hybrid RL ~\cite{jauhri2022reachability} use RL to condition base control on end-effector trajectories or 6D targets, while ReLMM focuses on autonomous real-world learning of modular navigation and grasping~\cite{sun2022relmm}.

\begin{figure}[t]
    \vspace{-4mm}
    \centering
    \includegraphics[width=0.94\linewidth]{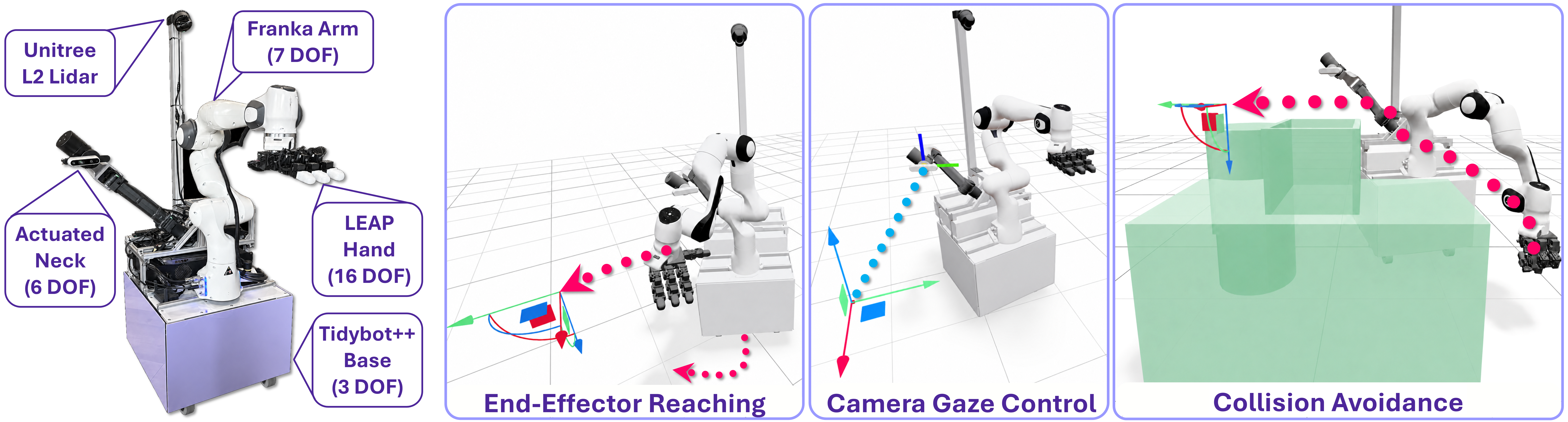}
    \vspace{-0.9em}
    \caption{
    \small \textbf{Left: Robot System.} Our platform consists of a holonomic TidyBot++~\cite{pmlr-v270-wu25a} base, a Franka arm, a LEAP Hand~\cite{shaw2023leaphand}, and an ARX X5 arm carrying a RealSense D435 RGB-D camera for active vision, totaling 32 degrees of freedom. The base is additionally equipped with a Unitree L2 3D LiDAR. 
    \textbf{Right: Vectorized WBC.} The vectorized fabrics WBC coordinates the base and arm to reach target end-effector poses, controls the active camera to track a look-at point, and performs self- and environment-collision avoidance.
    }
    \vspace{-6mm}
    \label{fig:real_system_and_fabrics}
\end{figure}

\vspace{-3mm}
\subsection{Sim-to-Real for Manipulation}
\vspace{-3mm}
Sim-to-real RL trains closed-loop manipulation policies in simulation for real-world deployment, enabling data scales beyond physical interaction alone. OpenAI Dactyl showed sim-to-real in-hand reorientation on a Shadow Hand via domain randomization~\cite{openaidactyl}. DeXtreme demonstrated agile Allegro-hand manipulation~\cite{dextreme}. Visual Dexterity reoriented novel object shapes from depth~\cite{visualdexterity}, and DexNDM improved in-hand rotation transfer with learned joint-wise dynamics~\cite{liu2026dexndm}. Together, these works show robust object- and dynamics-level dexterity but with only the reorientation task.

Beyond in-hand reorientation, recent systems train manipulation policies from different perceptual modalities, including point clouds~\cite{dexpoint,clutterdexgrasp, robustdexgrasp, dalal2025neural}, depth~\cite{dextrahg, dalal2024manipgen}, and RGB pixels~\cite{dextrahrgb, singh2025endtoendrlimprovesdexterous}. DexFunc and SimToolReal extend sim-trained control to affordance-guided grasps~\cite{dexfunc, simtoolreal}, while other work scales sim-to-real dexterity to higher-dimensional action spaces such as bimanual and humanoid manipulation~\cite{torulids,toruhumanoid}. These directions expand the complexity of policies trained in simulation, but are still largely evaluated in relatively fixed environments. In contrast, our framework scales visual policy training over joint object--environment variation.

ManipGen adopts a related local-policy formulation~\cite{dalal2024manipgen}, but relies on a separate inference-time motion planner. This assumption is difficult for mobile manipulation in unstructured scenes, where robust whole-body planning must operate directly from partial onboard observations. We instead distill local policies and whole-body motion generation into a single end-to-end policy for mobile manipulation, without requiring a whole-body motion planner at inference time.

\enlargethispage{1\baselineskip}
\vspace{-3mm}
\subsection{Whole-Body Control for Mobile Manipulation}
\label{sec:related_wbc}
\vspace{-3mm}
Whole-body control coordinates redundant degrees of freedom for task-space objectives like reaching, posture regulation, collision avoidance, and gaze control. QP-based differential IK libraries, such as Pink and Mink~\cite{pink,mink}, simplify composing tasks and constraints. While efficient for single-robot control, collision avoidance requires constrained solves with scene-dependent active constraints, making large-batch execution across many environments difficult.

Optimization-based WBC (e.g., PlaCo~\cite{placo}, OpenSoT~\cite{opensot}, ARC-OPT~\cite{arcopt}) and MPC frameworks like OCS2~\cite{ocs2} handle richer constraints and dynamics. However, they introduce per-step or horizon-dependent online optimization problems that are expensive to evaluate in batch. 

Reactive methods provide a different trade-off. RMPs produces expressive behavior through geometrically consistent pullbacks~\cite{rmp}, and geometric fabrics further compose reaching, posture regulation, collision avoidance, and gaze control as metric-weighted task-space policies~\cite{geometric_fabrics,acceleration_geometric_fabrics}.
We use fabrics not because they are universally more optimal than optimization-based WBC, but because they provide a fixed reactive computation graph that can be evaluated efficiently in parallel across many simulated environments.  Appendix~\ref{sec:wbc_tradeoffs} provides a more detailed comparison.

%===============================================================================
\newpage
\section{\method: Scaling Sim-to-Real for Contextual Whole-Body Manipulation}
\label{sec:method}
\vspace{-2mm}

We present \method, a sim-to-real framework for learning contextual whole-body manipulation at scale. Training RL manipulation policies directly across diverse scenes is difficult because policies must adapt their motion from high-dimensional scene observations, while rewards must remain effective across varied object--environment configurations. Rather than relying on RL to learn the full scene-dependent whole-body manipulation, we decompose the problem into \emph{global free-space reaching} and \emph{local contact-rich interaction}. Global reaching depends on scene context, including base placement, visibility, collision avoidance, but is largely free-space motion. To achieve this across arbitrary scenes, we introduce a vectorized environment-aware whole-body controller (WBC) in Sec.~\ref{sec:fabric_wbc} that brings the end effector near a target object while avoiding collisions, relieving RL from learning this scene-dependent behavior. 
Once the end effector is near the target, the remaining problem becomes a local contact-rich interaction, which can be learned by RL using compact local observations and simple rewards, detailed in Sec.~\ref{sec:local_rl}. \textit{This decomposition makes the COWBOY framework scalable:} a small set of local experts can be reused across thousands of full-scene layouts, while the vectorized WBC provides scene-aware whole-body approach behavior. Since both WBC and local RL experts require privileged scene information in simulation, the composed full teacher policies are distilled into a single vision-based whole-body policy Sec.~\ref{sec:distillation}. 
We demonstrate COWBOY on dexterous whole-body grasping. In Sec.~\ref{sec:results}, we deploy the resulting policy zero-shot in the real world and show that it generalizes across diverse objects and in-the-wild scenes.

% ===========================================
\vspace{-3mm}
\subsection{Vectorized Environment-Aware Whole-Body Controller}
\label{sec:fabric_wbc}
\vspace{-2mm}
We introduce a vectorized whole-body controller (WBC) that provides a compact interface for scene-aware whole-body motion across thousands of simulated environments. 
The WBC takes only an end-effector target pose and a 3D look-at point. 
Given an $SE(3)$ end-effector target, it coordinates the holonomic base and arm to reach the target while avoiding self-collisions and scene obstacles, without requiring an explicit base goal. 
Given a 3D look-at point, it controls the actuated neck to keep the onboard camera directed toward the object of interest. 
This interface serves two roles in our framework: it brings the end-effector from distant initial states to the local interaction region during distillation, and it provides a structured action abstraction for RL policies, allowing them to command task-space motion while the WBC handles whole-body coordination. 

We implement the WBC using geometric fabrics~\cite{fabrics_policy}, which enable behaviors such as collision avoidance while providing a fixed computation graph suitable for GPU-parallelized simulation. This avoids the per-environment constrained solves of QP-based IK and the horizon optimization of MPC; see Appendix~\ref{sec:wbc_tradeoffs}.
Fabrics compose task-space behaviors, such as reaching, posture regulation, and collision avoidance, through metric-weighted pullbacks.
Each term is defined by a task map $\mathbf{x}_i=\phi_i(\mathbf{q})$, velocity $\dot{\mathbf{x}}_i=\mathbf{J}_i(\mathbf{q})\dot{\mathbf{q}}$, metric $\mathbf{M}_i$, and force term $\mathbf{f}_i$, where $\mathbf{f}_i$ specifies the desired behavior and $\mathbf{M}_i$ sets its state-dependent priority.
All terms are pulled back and combined in joint space:
\vspace{-1mm}
\begin{equation}
    \left(\sum_i \mathbf{J}_i^\top \mathbf{M}_i \mathbf{J}_i\right)\ddot{\mathbf{q}}
    +
    \sum_i \mathbf{J}_i^\top
    \left(
        \mathbf{f}_i + \mathbf{M}_i \dot{\mathbf{J}}_i \dot{\mathbf{q}}
    \right)
    =
    \mathbf{0}.
    \label{eq:fabric_root}
\end{equation}
We then apply fabric energization and integrate the resulting acceleration $\mathbf{\ddot{q}}$ into base velocity and joint position commands. 
This formulation is naturally vectorizable: for $B$ environments, $\mathbf{q}\in\mathbb{R}^{B\times n}$, $\mathbf{J}_i\in\mathbb{R}^{B\times m_i\times n}$, and $\mathbf{M}_i\in\mathbb{R}^{B\times m_i\times m_i}$ are stored as batched tensors. Pullbacks $\mathbf{J}_i^\top \mathbf{M}_i \mathbf{J}_i$ are batched matrix multiplications, and combining fabric terms is a batched reduction. 

\textbf{Extending fabrics to mobile manipulators.} We implement the vectorized fabric WBC on the mobile manipulator shown in Fig.~\ref{fig:real_system_and_fabrics}. 
While prior work on fabrics primarily focuses on fixed-base arms \cite{geometric_fabrics, acceleration_geometric_fabrics}, we introduce fabric terms tailored for whole-body coordination, described below.

\enlargethispage{1\baselineskip}

\textbf{Whole-Body Reaching.}
To coordinate the base and arm given a single end-effector target, we employ two fabrics: \emph{base attractor} (BA) and \emph{end-effector attractor} (EEA). 
Both use the metric:
\begin{equation}
\mathbf{M}(\mathbf{x})
=
\eta\,\mathrm{diag}\!\left(
m_{\mathrm{far}}
+
\left(m_{\min}-m_{\mathrm{far}}\right)
\left[
\frac{1}{2}
\tanh\!\left(
-\alpha_m \left(\mathrm{dist}(\mathbf{x},\mathbf{x}^\star) - r_m\right)
\right)
+
\frac{1}{2}
\right]
\right),
\label{eq:metric_gate}
\end{equation}
where $r_m$ is the transition radius and $m_{\mathrm{far}},m_{\min}$ define the metric far from and near the target. 
For BA, $\mathbf{x}$ is the planar base pose and $\mathbf{x}^\star$ is derived from the planar projection of the end-effector target, with yaw facing the end-effector target. 
We set $m_{\min}=0$, so the base attractor smoothly vanishes once the target enters the arm workspace. 
For EEA, $\mathbf{x}$ is the end-effector pose and $\mathbf{x}^\star$ is the target pose; we set $m_{\mathrm{far}}=0$, so arm reaching activates only near the local manipulation workspace. This avoids premature arm extension while still allowing precise reaching; since the EEA Jacobian includes the full base-to-end-effector chain, the base is still actuated after BA vanishes.
Together, these terms produce coarse-to-fine whole-body reaching: the base handles distant motion, while arm motions are prioritized over the base when the robot is close to target for precise reaching.

\textbf{Gaze and Collision Avoidance.}
To alleviate the problem of partial observability in mobile manipulation, we add an $SO(3)$ camera-orientation attractor that points the camera optical axis toward the look-at point, together with a weak neck-posture regularizer. 
For collision avoidance, we approximate the robot with collision spheres and compute batched distances to other robot spheres and local environment primitives, including cuboids, capsules, and spheres. 
Each robot collision sphere induces a repulsion fabric that pushes it away from nearby obstacles or other robot spheres.

The full controller also includes posture regularization and joint-limit avoidance fabrics. 
Our implementation runs at $60$~Hz and supports up to $4096$ distinct simulation environments with an effective per-environment runtime of $0.03$~ms per control step. 
Additional details are provided in Appendix~\ref{sec:fabric_terms}.

% ===========================================
\vspace{-3mm}
\subsection{Contact-Rich Local RL Policies}
\label{sec:local_rl}
\vspace{-2mm}
COWBOY leverages the WBC for collision-free reaching motion across arbitrary environments, relieving RL from needing to learn this scene-dependent behavior.
Once the WBC brings the end effector near an object, the RL policy only needs to learn interactions with the object and reason about its immediate geometric constraints. Thus, we adopt the local policy formulation of~\cite{dalal2024manipgen}: policies are initialized near the target, observations are expressed in the end-effector frame, and actions are relative end-effector commands, executed by the WBC as well.

This local formulation has several advantages. Training in compact local scenes reduces simulation memory requirements; conditioning on local geometry rather than full-scene perception simplifies the observation space; and using the WBC as a structured action abstraction reduces action dimensionality.
Most importantly, because the policies are expressed in the end-effector frame, the same local behavior can be reused across many full-scene contexts during distillation. This quality makes COWBOY scalable across diverse environments. We discuss this more in Sec.~\ref{sec:distillation}.

\begin{figure}[t]
    \vspace{-6mm}
    \centering
    \includegraphics[width=0.95\linewidth]{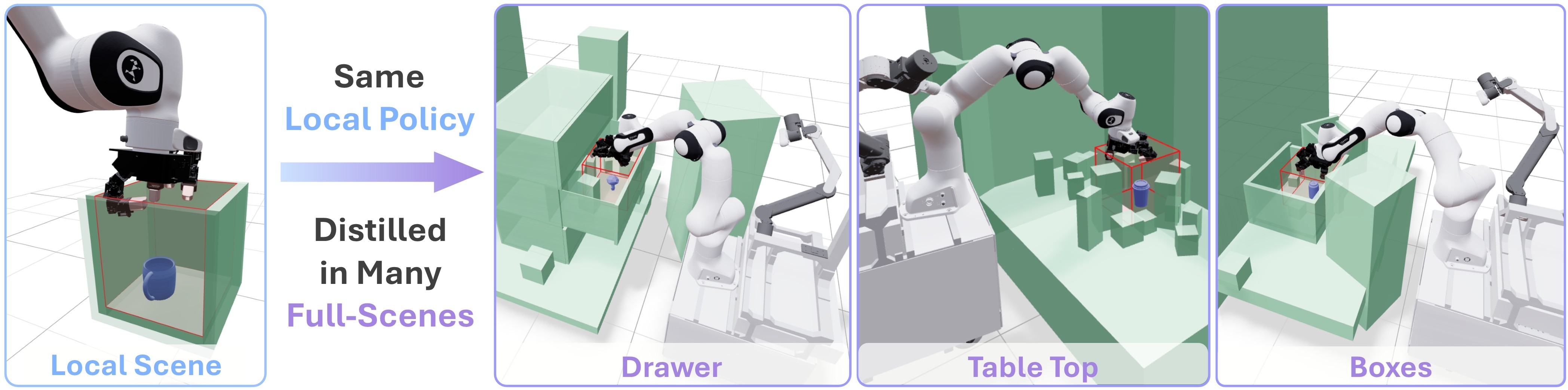}
    \vspace{-0.8em}
    \caption{ \small
    For a local scene, shown as the red-shaded volume, we train a local RL policy to manipulate the target object. We then embed the same local scene within diverse full scenes, where the WBC first brings the end effector near the object while avoiding collisions to match the local policy's initial state distribution. The local policy can then complete the task, enabling a policy trained in one local scene to generate expert motion across diverse full-scenes.
    }
    \vspace{-6mm}
    \label{fig:local_full_scenes}
\end{figure}

\textbf{Local Dexterous Grasping Experts.}
We instantiate this formulation for dexterous whole-body grasping by training three local RL experts, each corresponding to a distinct grasping affordance. Since each expert only needs to learn one type of grasp, reward specification is simple, facilitating RL training. Each expert is trained in a randomized cuboid local region with randomized target objects and wall configurations that encode spatial constraints. The \textbf{top-down grasp} expert operates in regions with side walls and an open top, illustrated in Fig.~\ref{fig:local_full_scenes} Left. The \textbf{side-grasp} expert approaches objects laterally, targeting elongated objects such as bottles. The \textbf{constrained side-grasp} expert also grasps laterally, but includes a top wall that requires the object to be extracted through a side opening, as in shelf-like settings. Local experts are visualized in Fig.~\ref{fig:local_policies_vis}. Additional manipulation skills, such as articulated-object interaction, can be incorporated by training additional local experts.

% \textbf{Local Grasping Experts.} For dexterous whole-body grasping, we train three local RL experts, each corresponding to a distinct grasping affordance. Each expert is trained in a randomized local cuboid region with randomized objects and wall configurations that encode spatial constraints. The \textbf{top-down grasp} expert operates in regions with side walls and an open top, illustrated in Fig.~\ref{fig:local_full_scenes} Left. The \textbf{side-grasp} expert approaches objects laterally, targeting elongated objects such as bottles. The \textbf{constrained side-grasp} expert also grasps laterally, but includes a top wall that requires the object to be extracted through a side opening, as in shelf-like settings.

\enlargethispage{1\baselineskip}

\textbf{Policy Formulation and Training.}
Each local policy receives privileged local observations: object bounding-box dimensions, local cuboid dimensions and wall configuration, LEAP Hand joint positions, and fingertip positions in the hand frame. 
The policy outputs relative end-effector commands and hand joint commands. 
Policies are trained in Isaac Gym~\cite{makoviychuk2021isaacgymhighperformance} using the RLgames~\cite{rl-games2021} implementation of PPO~\cite{schulman2017proximalpolicyoptimizationalgorithms}. 
Rewards and additional training details are provided in Appendix~\ref{sec:local_rl_details}.

% ===========================================
\vspace{-2 mm}
\subsection{Local-to-Full-Scenes Multi-Teacher Distillation}
\label{sec:distillation}
\vspace{-2mm}
We obtain a single real-world-deployable vision policy for contextual whole-body manipulation by distilling full-scene teachers composed of the environment-aware WBC from Sec.~\ref{sec:fabric_wbc} and the local RL experts from Sec.~\ref{sec:local_rl}. The distilled student performs contextual whole-body manipulation from onboard visual observations across environments and objects, no longer dependent on accurate full scene geometries and privileged object state information. Appendix~\ref{sec:distillation_details} provides full details.

\textbf{Full-Scene Environment Generation.}
For each environment, we sample a target object and local expert, then generate a full-scene layout containing a local interaction region compatible with that expert's training scene, as illustrated in Fig.~\ref{fig:local_full_scenes}. The expert type determines the valid scene family: top-down interaction regions occur in tabletop, drawer, or box-like scenes; side-grasp regions occur in scenes with elongated objects; and constrained side-grasp regions occur in shelf-like scenes requiring side extraction, visualized in Fig.~\ref{fig:full_scene_vis}. Since local experts are trained relative to the local interaction region, they can be reused in any full scene that preserves the same local geometric constraints. We randomize the surrounding scene geometry, clutter, and various physics parameters.

\textbf{Teacher Policy.}
Given a full scene and its compatible local expert, we compose the expert with the WBC to form a teacher policy. Each rollout begins with the robot initialized $0.2$--$2$~m from the target. During the \emph{approach phase}, the WBC brings the end effector to a staging pose near the local interaction region while avoiding obstacles and maintaining gaze on the target object. Once the end effector enters the local expert's initialization distribution, the \emph{interaction phase} begins: the expert outputs relative end-effector and hand commands, which are executed through the WBC. We also set the WBC look-at target to the object position such that the neck maintains gaze for active vision.

\begin{wrapfigure}{r}{0.51\textwidth}
    \vspace{-4mm}
    \centering
    \includegraphics[width=\linewidth]{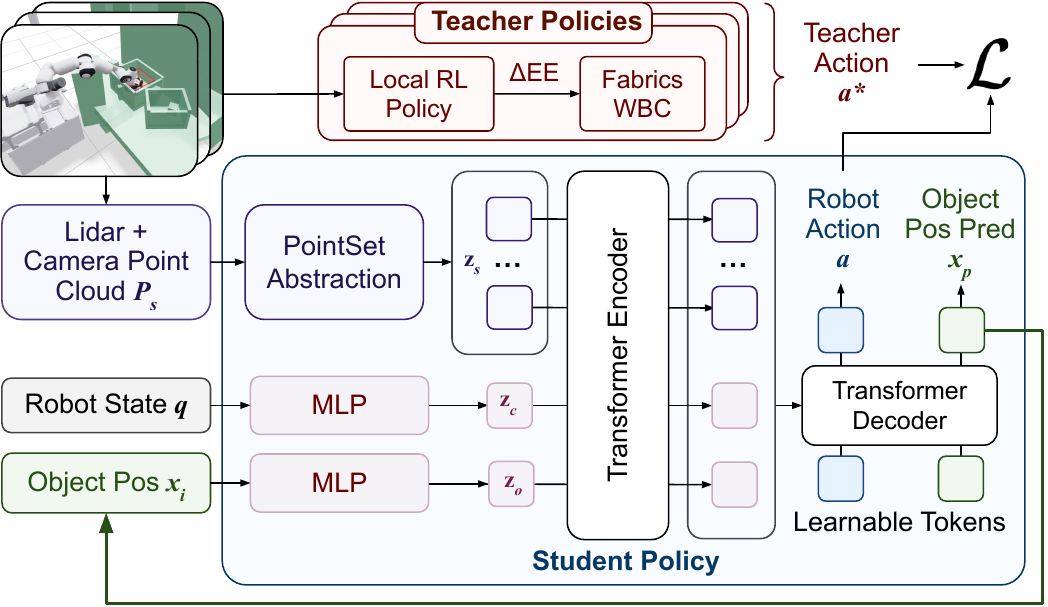}
    \vspace{-6mm}
    \caption{\small 
    We use a point-cloud conditioned transformer for the student policy, architecture inspired by \cite{drp}.
    }
    \label{fig:distillation}
    \vspace{-5mm}
\end{wrapfigure}

\textbf{Student Policy.}
We train a single onboard policy from the composed full-scene teachers using online DAgger~\cite{dagger}. During student rollouts, the teacher provides supervision for the current scene, while the student observes point clouds from the actuated-neck depth camera and base-mounted 3D LiDAR, expressed in the robot base frame. The student is a point-cloud-conditioned transformer \cite{drp}, illustrated in Fig.~\ref{fig:distillation}, and outputs base velocities and delta joint-position targets for the arm, hand, and actuated neck. Details are in Appendix~\ref{sec:student_policy_details}.

To scale distillation from partial onboard observations, we implement vectorized depth and LiDAR simulation by sampling points on robot and environment meshes and removing occluded points. Across $1024$ Isaac Gym environments on a single NVIDIA A6000, rendering depth observations at $212{\times}120$ resolution and LiDAR observations with $32768$ points takes $0.07$~s and $0.25$~s, respectively.

For whole-body dexterous grasping, the student is additionally conditioned on a 3D target-object position estimate, which specifies the desired object in cluttered scenes. Given an initial position target, the policy updates the estimate autoregressively using an auxiliary prediction head. Training targets are obtained from simulation and deployment targets can be initialized using a language-conditioned segmentation model such as SAM3~\cite{carion2025sam3segmentconcepts}. Details are in Appendix~\ref{sec:student_architecture_and_objective} and ~\ref{sec:sam3_localization}.

We distill three WBC--local-expert teacher families into one student across $8192$ full-scene environments and $250$ objects in Isaac Gym, distributed over $8$ NVIDIA A6000 GPUs. Additional training and sensor-simulation details are provided in Appendix~\ref{sec:student_architecture_and_objective} and \ref{sec:vectorized_sensor_rendering}. As shown in Sec.~\ref{sec:results}, in diverse real-world scenes, the resulting policy adapts its motion directly from observed object and scene geometry, without requiring explicit conditioning on base placement or grasp approach direction.

\newpage
\begin{figure}[t]
    \vspace{-3mm}
    \centering
    \includegraphics[width=1\linewidth]{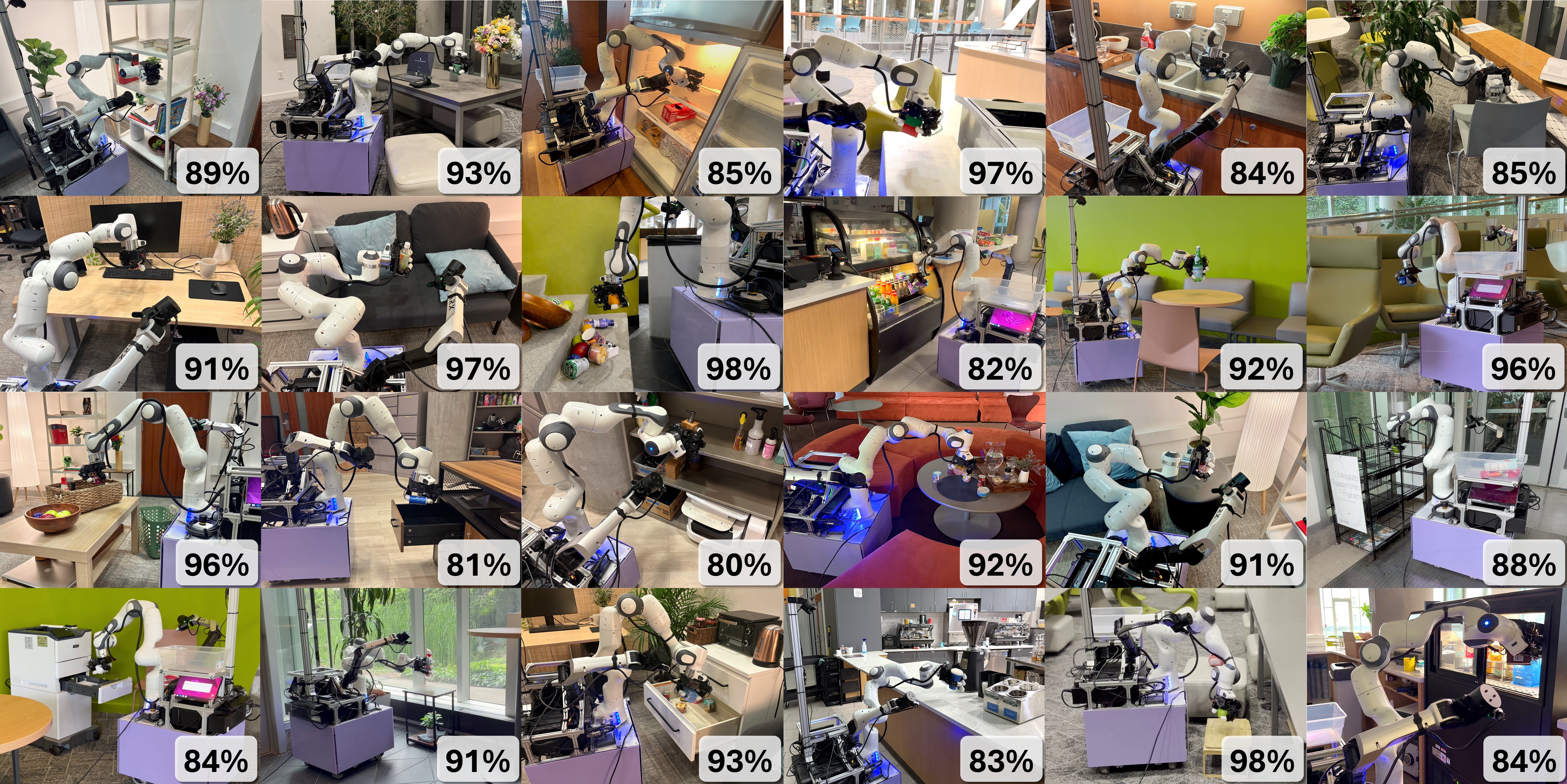}
    \vspace{-5mm}
    \caption{
    COWBOY achieves 91\% grasp success across 24 in-the-wild scenes and 100 objects. Bottom-right numbers show per-scene success over 25 objects with 3 trials each.
    }
    \vspace{-5mm}
    \label{fig:real_world_envs}
\end{figure}

\section{Experiments}
\begin{wrapfigure}{r}{0.46\textwidth}
    \vspace{-5mm}
    \centering
    \includegraphics[width=\linewidth]{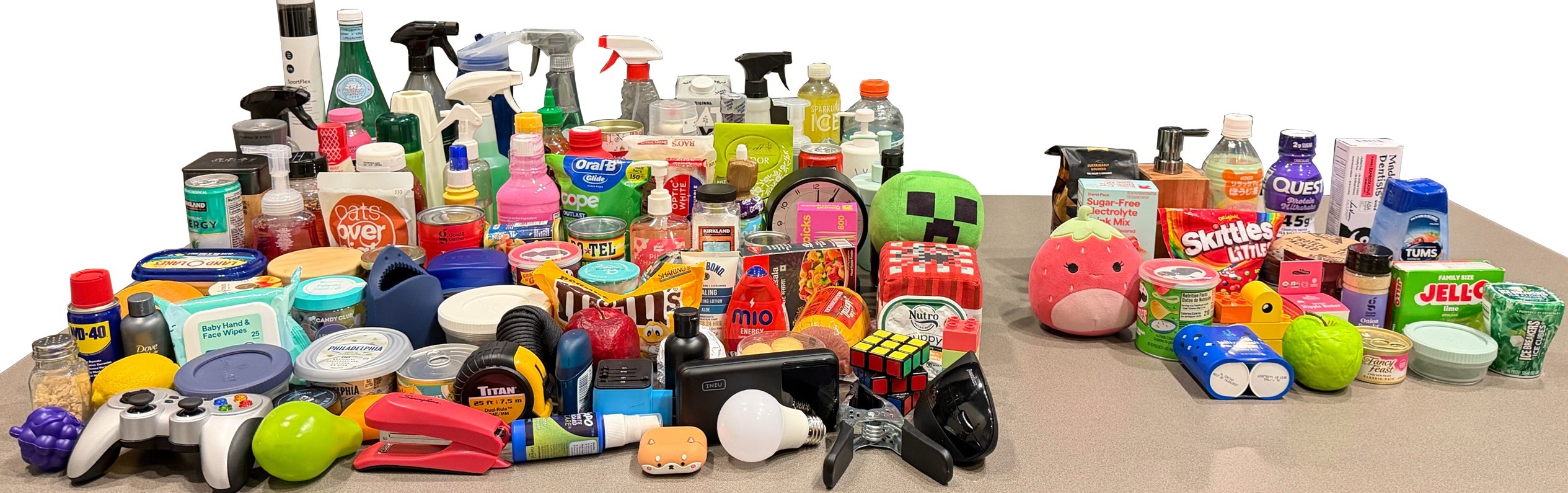}
    \vspace{-6mm}
    \caption{\small 
    Both sets of objects not seen at training.  \textbf{Left:} 100 random objects. \textbf{Right:} 20 fixed objects. 
    }
    \label{fig:real_object_set}
    \vspace{-4mm}
\end{wrapfigure}

\vspace{-3mm}
\textbf{Real-world evaluation.}
We evaluate the distilled dexterous whole-body grasping policy in 24 in-the-wild real-world scenarios, see Fig.~\ref{fig:real_world_envs}, including tabletops at different heights, shelves, boxes, chairs, fridges, drawers, and hybrid layouts that do not cleanly match any single simulated scene type mentioned in Sec.~\ref{sec:distillation}. For each scene, we evaluate on a fixed set of 20 objects. Using the same objects across scenes tests contextual whole-body manipulation. The same object should induce different base placements, viewpoints, and grasp strategies depending on the environment. To further test object generalization, we additionally sample 5 objects per scene from a random set of 100 objects. Both sets of objects are shown in Fig.~\ref{fig:real_object_set}. Each object per scene is evaluated under 3 orientations. A trial is successful if the object is lifted from initial pose by 10 cm without collisions.

\textbf{Real-world perception pipeline.}
We use SAM3~\cite{carion2025sam3segmentconcepts} to obtain an initial segmentation mask of the target object, which is lifted into 3D to estimate its position using the point-cloud centroid. During execution, the policy updates the target position autoregressively at 60 Hz through its auxiliary prediction head, while SAM3 provides periodic corrections at 3 Hz whenever a valid detection is available. This enables continuous target tracking between segmentation updates and during temporary detection dropouts. Additional details are provided in Appendix~\ref{sec:sam3_localization}.

\textbf{Simulation and ablations.}
We further evaluate COWBOY on 8192 held-out simulation environments and 150 held-out simulation objects, with results in Tab.~\ref{tab:sim_eval}. We ablate training with three variants of policy observation design: no active vision where the actuated neck is fixed at the default configuration, no LiDAR point clouds, and replacing rendered partial sensor observations with ground-truth full point clouds. We also ablate the WBC used for teacher policies by comparing our full fabric WBC against simpler variants in distillation environments, reporting grasp success, collision rate, manipulability, and joint-limit margin. Additional details regarding ablation experiments are provided in Appendix~\ref{sec:obs_ablation_details} and ~\ref{sec:wbc_ablation_details}.

\enlargethispage{1\baselineskip}
\vspace{-5mm}
\section{Results}
\label{sec:results}
\vspace{-4mm}
We organize results around six questions: (1) Does COWBOY generalize zero-shot to diverse real-world scenes and objects? (2) Does it produce contextual whole-body manipulation? (3) How should we design policy observations for sim-to-real mobile manipulation? (4) How does WBC design affect downstream teacher policy? (5) How much of the teacher policy performance tranfers to the student policy? (6) How does the student policy compare to other vision-based manipulation policies. Please see our video results at \url{cowboy-manip.com}

\textbf{COWBOY generalizes zero-shot to diverse real-world environments and objects.}
As seen in Fig.~\ref{fig:real_world_envs}, across 24 real-world scenes, COWBOY achieves a 91\% average grasp success rate despite large variation in scene geometry, object placement, and available approach directions.
These real scenes are never explicitly modeled in simulation during training: some, such as fridges, staircases, and chair scenes, are unlike any simulated scene type, while even tabletop- and shelf-like scenes are not explicitly modeled but only approximated through randomized procedural generation.
Performance is lower in scenes that require side grasps, such as shelves (see Fig.~\ref{fig:real_world_envs} top-left or bottom-right scenes), because side grasps risk knocking objects over in unrecoverable states.

COWBOY also succeeds on a broad range of everyday objects all unseen in simulation training as shown in Fig.~\ref{fig:real_object_set}, spanning diverse geometries as well as irregular and deformable objects. Overall, these results support our central claim: scaling to thousands of simulated object--environment configurations, enabled by environment-aware WBC and local-to-scene policy distillation, yields scene-conditioned behaviors that transfer zero-shot to diverse real-world settings.

\textbf{COWBOY policies exhibit contextual whole-body control.}
Beyond its high success rate, COWBOY adapts base motion, camera viewpoint, arm trajectories, and grasp strategies to both object and scene geometry, without needing explicit conditioning such as grasp type, base placement, or approach direction. 
As shown in the bottom right of Fig. \ref{fig:teaser}, in the same sofa scene, an elongated object like the bottle is grasped from the side, whereas a regularly shaped object like the red plushie is grasped from the top. In shelf scenes, where the surrounding geometry constrains the robot to a side grasp, the policy selects an appropriate lateral approach, as shown in top-left of Fig.~\ref{fig:teaser}. 

The policy also adapts whole-body motion for collision avoidance. As shown in Fig.~\ref{fig:teaser}, the base avoids obstacles such as the orange couch (bottom-left), compensates with arm reaching when base motion is blocked such as by the gray stool (top-right), and avoids sides of basket when reaching in.  These behaviors show that the policy is not replaying a fixed grasp primitive but instead understanding the full scene context. This context-aware whole-body behavior emerges from distilling environment-aware WBC and local experts across diverse simulated scenes and objects.

\begin{table*}[t]
\vspace{-6mm}
\centering
\scriptsize
\setlength{\tabcolsep}{2.4pt}
\renewcommand{\arraystretch}{1.06}

\begin{minipage}[t]{0.39\textwidth}
\vspace{0pt}
\centering
\begin{tabularx}{\linewidth}{@{}>{\raggedright\arraybackslash}p{0.35\linewidth}@{}*{5}{Y}@{}}
\toprule
& \multicolumn{5}{c}{\makecell{\textbf{Grasp Success \%} $\uparrow$}} \\
\cmidrule(lr){2-6}
\textbf{Policy}
& \makecell{\tiny\textbf{Table}\\[-1pt]\tiny\textbf{top}}
& {\tiny\textbf{Drawer}}
& {\tiny\textbf{Box}}
& {\tiny\textbf{Shelf}}
& {\tiny\textbf{Avg.}} \\
\midrule
End-to-end RL
& \bad{0.0} & \bad{0.0} & \bad{0.0} & \bad{0.0} & \bad{0.0} \\
ManipGen~\cite{dalal2024manipgen}
& \bad{54.8} & \midperf{89.2} & \midperf{87.2}
& \bad{21.2} & \midperf{63.1} \\
\textbf{COWBOY}
& \good{94.7} & \good{92.9} & \good{91.3}
& \good{90.5} & \good{92.4} \\
\bottomrule
\end{tabularx}
\vspace{-3mm}
\caption{\small Comparison with learning baselines on held-out simulation environments.}
\label{tab:baseline_comparison}
\end{minipage}
\hfill
\begin{minipage}[t]{0.59\textwidth}
\vspace{0pt}
\centering
\begin{tabularx}{\linewidth}{@{}>{\raggedright\arraybackslash}p{0.18\linewidth}@{}*{10}{Y}@{}}
\toprule
& \multicolumn{5}{c}{\makecell{\textbf{Grasp Success \%} $\uparrow$}}
& \multicolumn{5}{c}{\makecell{\textbf{Collision \%} $\downarrow$}} \\
\cmidrule(lr){2-6}\cmidrule(l){7-11}
\textbf{Policy}
& \makecell{\tiny\textbf{Table}\\[-1pt]\tiny\textbf{top}}
& {\tiny\textbf{Drawer}}
& {\tiny\textbf{Box}}
& {\tiny\textbf{Shelf}}
& {\tiny\textbf{Avg.}}
& \makecell{\tiny\textbf{Table}\\[-1pt]\tiny\textbf{top}}
& {\tiny\textbf{Drawer}}
& {\tiny\textbf{Box}}
& {\tiny\textbf{Shelf}}
& {\tiny\textbf{Avg.}} \\
\midrule
Teacher
& 96.0 & 95.6 & 93.4 & 97.3 & 95.6
& 3.9 & 4.8 & 6.2 & 2.4 & 4.3 \\
Student
& 94.7 & 92.9 & 91.3 & 90.5 & 92.4
& 5.3 & 7.2 & 8.1 & 3.6 & 6.1 \\
\bottomrule
\end{tabularx}
\vspace{-3mm}
\caption{\small Held-out simulation evaluation across $8{,}000$ environments and $150$ objects. The student closely matches the teacher across scene categories.}
\label{tab:sim_eval}
\end{minipage}

\vspace{-6mm}
\end{table*}

\textbf{COWBOY achieves higher grasp success compared to alternative vision-based policies.}
\label{sec:external_abalations}
We evaluate our student policy against an end-to-end RL policy and ManipGen \cite{dalal2024manipgen} on the same held out simulation environments in Tab.~\ref{tab:sim_eval} under the same metrics. The point-cloud conditioned end-to-end RL achieves 0\% despite varying rewards, curricula, and visual encoders. As a sanity check, without obstacles in simulation environments, end-to-end RL achieves 96\%, highlighting the difficulty of jointly learning scene reasoning, collision avoidance, and dexterous interaction. ManipGen uses point-cloud-conditioned local policies, AIT* motion planning, and a VLM to specify a planning target \cite{dalal2024manipgen}. This design creates a bottleneck for intermediate-goal selection: the VLM-specified planning target does not account for downstream manipulation and often leaves the arm near singularities. COWBOY instead distills reaching and manipulation jointly into a single policy, removing this intermediate decision, achieving higher performance on all four scene categories.

It is also important to note that ManipGen trains thousands of single object experts, while COWBOY trains every expert across hundreds of objects. The tradeoff is that only a few experts can supervise thousands of full scene-object contexts.

\textbf{Teacher performance transfer.}
\label{sec:sim_evals}
We evaluate both teacher and student policies on $8{,}000$ held-out simulation environments spanning four scene categories: tabletops, drawers, boxes, and shelves. We additionally evaluate on $150$ held-out objects not used during training. A grasp is successful if the object is lifted at least $10$~cm from its initial position. The WBC can enter the local RL expert distribution in 98.4\% of trials, with failures included in the teacher success rate ablations. Collision rate is the percentage of environments with either robot-environment collision or self-collision, excluding intentional contact with the target object.

The student closely matches teacher performance in tabletops, drawers, and boxes, suggesting that the distilled vision policy captures the scene-conditioned behaviors generated by the WBC and local experts. Performance drops more noticeably in shelves, likely due to heavier occlusion and the need for more precise obstacle-aware motion during side extraction. Nevertheless, the student maintains low collision rates across all scene types, indicating that the distilled policy preserves much of the teacher's scene-aware whole-body behavior.

\begin{table*}[t]
\vspace{-6mm}
\centering
\scriptsize
\setlength{\tabcolsep}{2.4pt}
\renewcommand{\arraystretch}{1.06}

\begin{minipage}[t]{0.42\textwidth}
\vspace{0pt}
\centering
\begin{tabularx}{\linewidth}{@{}>{\raggedright\arraybackslash}p{0.31\linewidth}@{}Y@{}Y@{}Y@{}Y@{}}
\toprule
& \multicolumn{2}{c}{\makecell{\textbf{Grasp Success \%} $\uparrow$}} 
& \multicolumn{2}{c}{\makecell{\textbf{Collision \%} $\downarrow$}} \\
\cmidrule(lr){2-3}\cmidrule(l){4-5}
\textbf{Variant} & \textbf{Sim} & \textbf{Real} & \textbf{Sim} & \textbf{Real} \\
\midrule
No active vision & \midperf{78.8} & \bad{46.7} & \midperf{9.4} & \good{8.2} \\
No LiDAR input   & \midperf{75.2} & \midperf{71.1} & \bad{16.7} & \bad{31.7} \\
No sim. render   & \good{94.4} & \bad{0.0} & \good{5.6} & \midperf{11.3} \\
\textbf{Ours}    & \good{92.4} & \good{91.1} & \good{6.4} & \good{3.9} \\
\bottomrule
\end{tabularx}
\vspace{-3mm}
\caption{\small Student policy observation ablation.}
\label{tab:observation_ablation}
\end{minipage}
\hfill
\begin{minipage}[t]{0.55\textwidth}
\vspace{0pt}
\centering
\begin{tabularx}{\linewidth}{@{}>{\raggedright\arraybackslash}p{0.37\linewidth}@{}Y@{}Y@{}Y@{}Y@{}}
\toprule
\textbf{Controller} 
& \begin{tabular}[b]{@{}c@{}}\textbf{Grasp}\\\textbf{Success \%} $\uparrow$\end{tabular}
& \begin{tabular}[b]{@{}c@{}}\textbf{Collision}\\\textbf{\%} $\downarrow$\end{tabular}
& \begin{tabular}[b]{@{}c@{}}\textbf{Arm}\\\textbf{Manip.} $\uparrow$\end{tabular}
& \begin{tabular}[b]{@{}c@{}}\textbf{Joint-Limit}\\\textbf{Margin} $\uparrow$\end{tabular} \\
\midrule
Weighted IK 
& \bad{43.6} 
& \bad{41.4} 
& \bad{0.039} 
& \bad{1.51} \\
\makecell[l]{Fabrics WBC without\\collision avoidance} 
& \midperf{74.8} 
& \bad{32.6} 
& \good{0.067} 
& \midperf{1.87} \\
\textbf{Full Fabrics WBC (Ours)} 
& \good{95.6} 
& \good{4.3} 
& \good{0.072} 
& \good{1.99} \\
\bottomrule
\end{tabularx}
\vspace{-3mm}
\caption{\small We ablate WBC on teacher grasp success, collision rate, manipulability, and joint-limit margin (radians).}
\label{tab:wbc_ablation}
\end{minipage}
\vspace{-6mm}
\end{table*}

\textbf{Effect of policy observation design.}
Tab.~\ref{tab:observation_ablation} shows that each observation component is critical for real-world grasping. Active vision improves success by keeping near-ground objects and box scenes in view, expanding the observable workspace beyond a fixed neck configuration. LiDAR reduces collisions by providing global scene context often missed by the actuated neck depth camera due to limited field of view and minimum sensing distance. Without LiDAR point clouds, the robot often collides with nearby structures such as the table. Finally, training with ground-truth full point clouds performs well in simulation but fails in the real world, showing that rendered partial sensor observations are necessary for sim-to-real transfer.

\textbf{Effect of WBC design.}
Tab.~\ref{tab:wbc_ablation} shows that the WBC used by teacher policies during both reaching and interaction phases strongly affects downstream teacher policy performance.
Weighted IK often stretches the arm toward distant targets instead of using the base for approach, leading to low manipulability, causing joints at their joint limits, frequent collisions, and subsequently poor grasp success.
Fabric WBC without collision avoidance still suffers from frequent collisions, but its whole-body reaching terms, BA and EEA, improve base-arm coordination and prevent premature arm extension, leading to higher manipulability and larger joint-limit margins.
The full fabric WBC achieves the highest grasp success and lowest collision rate while preserving strong manipulability and joint-limit margins, demonstrating its value for learning scene-aware whole-body mobile manipulation.

\enlargethispage{1\baselineskip}

\vspace{-4mm}
\section{Conclusion and Limitations}
\vspace{-4mm}
We present COWBOY, a scalable sim-to-real framework for contextual whole-body manipulation that combines a vectorized environment-aware WBC, local RL interaction experts, and full-scene distillation. COWBOY enables mobile manipulation training across thousands of simulated object--environment configurations, resulting in whole-body policies that transfer zero-shot to diverse in-the-wild scenes. The learned grasping policy adapts base motion, camera viewpoint, and grasp strategy from observed object and scene geometry, displaying contextual whole-body manipulation.

\vspace{-1mm}
\textbf{Limitations.}
For our whole-body grasping policy, the grasp strategy is inferred from scene and object context, but many objects admit multiple valid grasps depending on the downstream use case. Functional whole-body grasping therefore requires additional conditioning beyond scene alone.
Nonetheless, COWBOY provides a scalable framework for training sim-to-real policies that use object and scene context for whole-body manipulation.

\acknowledgments{}
We would like to thank Tony Tao, Hengkai Pan, Sungjae Park, Ritvik Singh, Arthur Allshire, Tal Daniel, Haibo Zhao, Arth Shukla and Dieter Fox for valuable discussions and feedback on this work. We are also grateful to Adam Kan for his assistance in setting up the TidyBot++ base, and to Peiqi Liu and Sri Anumakonda for their help with experiments. This work was supported in part by ONR N000142312368 grant, ONR MURI N00014-25-1-2116, AFOSR FA9550-23-1-0747, ONR MURI N00014-22-1-2773 and ONR MURI N00014-24-1-2748.

%===============================================================================

\clearpage
% % The acknowledgments are automatically included only in the final and preprint versions of the paper.
% \acknowledgments{If a paper is accepted, the final camera-ready version will (and probably should) include acknowledgments. All acknowledgments go at the end of the paper, including thanks to reviewers who gave useful comments, to colleagues who contributed to the ideas, and to funding agencies and corporate sponsors that provided financial support.}

%===============================================================================

\bibliography{references}  % .bib

\newpage
\appendix
\input{appendix}

\end{document}

%% file: appendix.tex
% ============================================================
% Appendix: WBC Trade-Offs
% ============================================================
\section*{Appendix}
\section{Whole-Body Controller Implementation Trade-offs}
\label{sec:wbc_tradeoffs}
This section expands Sec.~\ref{sec:related_wbc}. We do not claim that geometric fabrics are universally superior to optimization-based WBC; rather, fabrics provide the trade-off we need for large-scale RL: reactive collision-aware control with a fixed batched computation graph.

\textbf{Representative implementations.}
Table~\ref{tab:wbc_implementation_tradeoffs} compares public controller implementations along properties relevant to our setting. ``Native Batch API'' indicates whether the implementation exposes an interface intended to evaluate the controller across thousands of independent robot-scene instances. This is an implementation-level comparison, not a theoretical limitation or runtime benchmark.

\begin{table}[h]
\centering
\small
\setlength{\tabcolsep}{4pt}
\begin{tabular}{lcccc}
\toprule
\textbf{Implementation} &
\makecell{\textbf{Controller} \\ \textbf{Family}} &
\makecell{\textbf{Collision} \\ \textbf{Avoidance}} &
\makecell{\textbf{Horizon} \\ \textbf{Based}} &
\makecell{\textbf{Native Batch} \\ \textbf{API}} \\
\midrule
Pink~\cite{pink}        & Differential IK / QP IK        & Partial & No  & No \\
Mink~\cite{mink}        & Differential IK / QP IK        & Yes     & No  & No \\
PlaCo~\cite{placo}      & QP-based whole-body IK/dynamics & Yes    & No  & No \\
OpenSoT~\cite{opensot}  & Hierarchical QP WBC             & Yes     & No  & No \\
ARC-OPT WBC~\cite{arcopt} & Optimization-based WBC        & Yes     & No  & No \\
OCS2~\cite{ocs2}        & MPC / optimal control           & Yes     & Yes & No \\
Ours                    & Geometric fabrics               & Yes     & No  & Yes \\
\bottomrule
\end{tabular}
\vspace{1mm}
\caption{
Representative public controller implementations compared along dimensions relevant to large-scale robot learning.
The table is not a runtime benchmark; it motivates our use of a fixed-graph geometric-fabric WBC for scalable RL.
}
\label{tab:wbc_implementation_tradeoffs}
\end{table}

\vspace{-3mm}
\textbf{Differential IK and QP-based IK.}
\label{sec:weighted_ik}
Differential IK computes joint velocities that locally reduce task-space error, typically through a least-squares objective:
\begin{equation*}
\min_{\dot{\mathbf{q}}}
\left\|
\mathbf{J}_{\mathrm{ee}}(\mathbf{q})\dot{\mathbf{q}} - \dot{\mathbf{x}}_{\mathrm{ee}}^*
\right\|^2
+
\lambda \|\dot{\mathbf{q}}\|^2 .
\end{equation*}
In this unconstrained form, the computation is local and easy to batch. Practical IK libraries, however, commonly add joint limits, velocity limits, and collision avoidance through QP constraints. For example, Pink formulates task-based IK with weighted tasks and limits~\cite{pink}, and Mink solves a per-control-step QP over joint velocity while respecting tasks, limits, and collision constraints~\cite{mink}. A typical constrained IK problem is:
\begin{align*}
\min_{\dot{\mathbf{q}}} \quad
& \left\|
\mathbf{J}_{\mathrm{ee}}(\mathbf{q})\dot{\mathbf{q}} - \dot{\mathbf{x}}_{\mathrm{ee}}^*
\right\|^2
+
\lambda \|\dot{\mathbf{q}}\|^2 \\
\text{s.t.} \quad
& \dot{\mathbf{q}}_{\min} \leq \dot{\mathbf{q}} \leq \dot{\mathbf{q}}_{\max}, \\
& \nabla d_j(\mathbf{q})^\top \dot{\mathbf{q}}
\geq
-\alpha\left(d_j(\mathbf{q})-d_{\mathrm{safe}}\right),
\quad j=1,\ldots,N_c .
\end{align*}
Here $d_j(\mathbf{q})$ is a robot-robot or robot-obstacle distance. Such QPs are effective for a single robot, but in our setting thousands of environments have different robot states, obstacle layouts, and active constraints. Although batched QP solvers exist in principle, the computation is less naturally expressed as one fixed tensor program than the fabric pullback computation.

\textbf{Optimization-based WBC.}
Optimization-based WBC generalizes constrained IK to richer whole-body tasks. PlaCo provides a QP framework where tasks and constraints are specified in task space~\cite{placo}; OpenSoT provides hierarchical whole-body control for robots subject to constraints~\cite{opensot}; and ARC-OPT provides velocity-, acceleration-, and torque-level WBC implementations based on costs and constraints~\cite{arcopt}. These frameworks are powerful for single-robot WBC, but they still rely on per-instance optimization solves at each control step, which is not the computation pattern we want for thousands of parallel RL environments.

\textbf{MPC and optimal control.}
MPC methods, such as OCS2, optimize future trajectories subject to dynamics, costs, and constraints~\cite{ocs2}:
\begin{align*}
\min_{\{\mathbf{x}_k,\mathbf{u}_k\}_{k=0}^{H}} \quad
& \sum_{k=0}^{H-1} \ell(\mathbf{x}_k,\mathbf{u}_k) + \ell_f(\mathbf{x}_H) \\
\text{s.t.} \quad
& \mathbf{x}_0 = \mathbf{x}_{\mathrm{current}}, \\
& \mathbf{x}_{k+1} = f(\mathbf{x}_k,\mathbf{u}_k), \\
& g(\mathbf{x}_k,\mathbf{u}_k) \leq 0,
\quad k=0,\ldots,H-1 .
\end{align*}
Here $H$ is the prediction horizon, $\mathbf{x}_k\in\mathbb{R}^{n_x}$ is the system state at horizon step $k$, and $\mathbf{u}_k\in\mathbb{R}^{n_u}$ is the control input, where $n_x$ and $n_u$ denote the state and control dimensions, respectively.
In a multiple-shooting formulation, the number of optimization variables scales as $O(H(n_x+n_u))$, since the solver optimizes over both predicted states and controls.
Collision constraints are also replicated across horizon steps: if the robot has $N_s$ collision primitives and the local scene contains $N_o$ obstacles, enforcing all pairwise distance constraints at each step can require $O(HN_sN_o)$ collision constraints.
This horizon-based formulation is powerful, but the resulting per-environment optimization problem becomes expensive when solved independently across thousands of high-DoF mobile-manipulation environments.

\textbf{Why geometric fabrics.}
Geometric fabrics provide a reactive alternative. They define local second-order policies in task spaces and combine them through metric-weighted pullbacks~\cite{geometric_fabrics}:
\begin{equation*}
(\mathbf{q},\dot{\mathbf{q}})
\rightarrow
\{\phi_i(\mathbf{q}), \mathbf{J}_i(\mathbf{q}), \dot{\mathbf{J}}_i(\mathbf{q},\dot{\mathbf{q}})\}_i
\rightarrow
\{\mathbf{M}_i, \mathbf{f}_i\}_i
\rightarrow
\sum_i \left(
\mathbf{J}_i^\top \mathbf{M}_i \mathbf{J}_i,\;
\mathbf{J}_i^\top(\mathbf{f}_i + \mathbf{M}_i\dot{\mathbf{J}}_i\dot{\mathbf{q}})
\right)
\rightarrow
\ddot{\mathbf{q}} .
\end{equation*}
This fixed structure is GPU-friendly: task evaluations are independent across environments, pullbacks are batched matrix multiplications, and the root solve has identical shape across the batch. This makes fabrics a natural fit for large-scale RL rollouts while still supporting reaching, collision avoidance, posture regulation, and gaze control.

% ============================================================
% Appendix: Fabric Formulation and Terms
% ============================================================
\section{Geometric Fabrics Preliminaries}
\label{sec:fabric_prelim}

This section provides additional details for the geometric-fabric formulation \cite{geometric_fabrics} used by our vectorized environment-aware WBC.
Fabrics provide a framework for specifying second-order behaviors in simple task spaces and composing them into a single configuration-space dynamical system that is covariant under task transformations.

\subsection{Motion Policies and Fabric Composition}

A motion policy is a second-order vector field on a task manifold $\mathcal{X}$,
\begin{equation*}
    \ddot{\mathbf{x}} = \boldsymbol{\pi}(\mathbf{x},\dot{\mathbf{x}}).
\end{equation*}
Fabrics compose many such task-space policies into root dynamics on the robot configuration manifold $\mathcal{Q}$.
Let $\mathbf{q}\in\mathcal{Q}$ and $\dot{\mathbf{q}}\in T_{\mathbf{q}}\mathcal{Q}$.
Each task is defined by a differentiable map
\begin{equation*}
    \mathbf{x}_i=\phi_i(\mathbf{q}), 
    \qquad 
    \dot{\mathbf{x}}_i=\mathbf{J}_i(\mathbf{q})\dot{\mathbf{q}}, 
    \qquad 
    \mathbf{J}_i=\frac{\partial \phi_i}{\partial \mathbf{q}} .
\end{equation*}

Following the notation in Sec.~\ref{sec:fabric_wbc}, each fabric term is represented by a task-space metric $\mathbf{M}_i$ and force term $\mathbf{f}_i$:
\begin{equation*}
    \mathcal{F}_i = (\mathbf{M}_i,\mathbf{f}_i),
\end{equation*}
where $\mathbf{f}_i$ specifies the desired task-space behavior and $\mathbf{M}_i\succeq 0$ specifies the state-dependent priority and geometry of that behavior.

At each control step, all task-space terms are pulled back to the root configuration space and combined:
\begin{equation*}
    \left(\sum_i \mathbf{J}_i^\top \mathbf{M}_i \mathbf{J}_i\right)\ddot{\mathbf{q}}
    +
    \sum_i \mathbf{J}_i^\top
    \left(
        \mathbf{f}_i + \mathbf{M}_i \dot{\mathbf{J}}_i \dot{\mathbf{q}}
    \right)
    =
    \mathbf{0}.
    \label{eq:app_fabric_root}
\end{equation*}
Equivalently, defining
\begin{equation*}
    \mathbf{M}(\mathbf{q},\dot{\mathbf{q}})
    =
    \sum_i \mathbf{J}_i^\top \mathbf{M}_i \mathbf{J}_i
    \qquad
    \mathbf{f}(\mathbf{q},\dot{\mathbf{q}})
    =
    \sum_i \mathbf{J}_i^\top
    \left(
        \mathbf{f}_i + \mathbf{M}_i \dot{\mathbf{J}}_i \dot{\mathbf{q}}
    \right),
\end{equation*}
the raw composed acceleration is
\begin{equation*}
    \ddot{\mathbf{q}}_{\mathrm{raw}}
    =
    -\mathbf{M}(\mathbf{q},\dot{\mathbf{q}})^{-1}
    \mathbf{f}(\mathbf{q},\dot{\mathbf{q}}).
\end{equation*}

We then apply the standard fabric energization step,
\begin{equation*}
    \ddot{\mathbf{q}}
    =
    \mathrm{Energize}
    \left(
        \ddot{\mathbf{q}}_{\mathrm{raw}},
        \mathbf{M},
        \mathbf{q},
        \dot{\mathbf{q}}
    \right),
\end{equation*}
which modifies the acceleration along the velocity direction so that the dynamics are compatible with a chosen energy function.
This separates path geometry from speed regulation, allowing diverse local behaviors to be composed while preserving the stability properties of geometric fabrics.

For batched simulation, all quantities are stored across environments. Here \(B\) denotes the number of parallel environments, \(n=\dim(\mathcal{Q})\) is the robot configuration dimension, and \(m_i=\dim(\mathcal{X}_i)\) is the dimension of the \(i\)-th task space:
\begin{equation*}
    \mathbf{q}\in\mathbb{R}^{B\times n},
    \qquad
    \mathbf{J}_i\in\mathbb{R}^{B\times m_i\times n},
    \qquad
    \mathbf{M}_i\in\mathbb{R}^{B\times m_i\times m_i}.
\end{equation*}
Thus, pullbacks such as $\mathbf{J}_i^\top\mathbf{M}_i\mathbf{J}_i$ are batched matrix multiplications, and combining fabric terms is a batched reduction.
We next describe the fabric components used in our WBC.

\section{Vectorized Environment-Aware Fabric WBC Details}
\label{sec:fabric_terms}

This section details our vectorized environment-aware fabric WBC. The controller exposes a compact interface for whole-body mobile manipulation. Given a single end-effector target pose, the WBC coordinates the base and arm to reach the target while avoiding collisions, without requiring an additional base target or posture command.
This simplified interface is useful for both scene-aware reaching during distillation and as a compact action abstraction for vectorized local RL policies.

Our WBC instantiates several fabric terms.
For whole-body coordination, we use an \emph{end-effector attractor} (EEA), a \emph{base attractor} (BA), and \emph{posture regularization} (PR).
We also use \emph{joint-limit repulsion} (JLR) on both the Franka arm and ARX camera arm to keep joints away from their limits.
For collision avoidance, we use \emph{body repulsion} (BR) defined on distances between robot collision spheres and nearby obstacles or robot links.
For active vision, we use a \emph{gaze control attractor} (GCA) to orient the actuated camera toward a specified look-at point. Only limited tuning is required to produce stable WBC behavior. We use the same parameter set across all training tasks without per-scene tuning.

\subsection{Fabrics WBC Configuration Space and Interface}

The WBC operates on the mobile-manipulator configuration
\[
\mathbf{q}
=
\left[
x_{\mathrm{base}},
y_{\mathrm{base}},
\theta_{\mathrm{base}},
\mathbf{q}_{\mathrm{Franka}},
\mathbf{q}_{\mathrm{ARX}}
\right],
\]
where the holonomic base is represented by a PPR parameterization: two prismatic coordinates for world-frame translation and one revolute coordinate for yaw.
The Franka arm is used for manipulation, and the ARX arm carries the onboard camera for active vision.
The LEAP hand is not included in the fabric state; hand commands are provided by a learned manipulation policy directly.

The WBC receives
\[
\mathbf{T}^{\star}_{\mathrm{ee}}\in SE(3),
\qquad
\mathbf{p}^{\star}_{\mathrm{gaze}}\in\mathbb{R}^3,
\qquad
\mathcal{O},
\]
where $\mathbf{T}^{\star}_{\mathrm{ee}}$ is the desired end-effector pose, $\mathbf{p}^{\star}_{\mathrm{gaze}}$ is the 3D look-at point for the camera, and $\mathcal{O}$ denotes local environment geometry used for collision avoidance.
The WBC outputs base velocity and joint-position commands after composing all fabric terms, applying energization, and integrating the resulting acceleration.

\subsection{End-Effector Attractor (EEA)}

The EEA drives the manipulation end-effector toward the target pose $\mathbf{T}^{\star}_{\mathrm{ee}}$.
Following the keypoint pose representation used in geometric fabrics~\cite{geometric_fabrics}, we represent an $SE(3)$ pose as seven 3D keypoints: one at the frame origin and six axis-aligned offset points.
Let
\[
\mathcal{R}
=
\left\{
\mathbf{0},
\pm \delta\mathbf{e}_x,
\pm \delta\mathbf{e}_y,
\pm \delta\mathbf{e}_z
\right\},
\qquad
\delta = 0.25~\mathrm{m}.
\]
For a pose $\mathbf{T}=(\mathbf{R},\mathbf{p})$, define
\[
\kappa(\mathbf{T})
=
\left[
\mathbf{p}+\mathbf{R}\mathbf{r}_1;
\ldots;
\mathbf{p}+\mathbf{R}\mathbf{r}_7
\right]
\in \mathbb{R}^{21},
\qquad
\mathbf{r}_j\in\mathcal{R}.
\]
The EEA task variable and target are
\[
\mathbf{x}^{\mathrm{eef}}
=
\kappa(\mathbf{T}_{\mathrm{ee}}(\mathbf{q})),
\qquad
\mathbf{x}^{\mathrm{eef}\star}
=
\kappa(\mathbf{T}^{\star}_{\mathrm{ee}}).
\]

The EEA uses an attractor force that drives $\mathbf{x}^{\mathrm{eef}}$ toward $\mathbf{x}^{\mathrm{eef}\star}$.
With
\[
\mathbf{e}
=
\mathbf{x}^{\mathrm{eef}}-\mathbf{x}^{\mathrm{eef}\star},
\qquad
\rho=\|\mathbf{e}\|,
\]
the task-space acceleration policy is
\[
\mathbf{a}_{\mathrm{att}}
=
-k\,\tanh(\alpha_f \rho)
\frac{\mathbf{e}}{\rho+\epsilon}
-
d(\rho)\dot{\mathbf{x}}^{\mathrm{eef}},
\qquad
\mathbf{f}_{\mathrm{EEA}}
=
-\mathbf{M}_{\mathrm{EEA}}\mathbf{a}_{\mathrm{att}}.
\]

To avoid stretching the arm toward unreachable targets, the EEA metric is activated only when the target enters the arm's local workspace:
\[
\mathbf{M}_{\mathrm{EEA}}
=
\eta_{\mathrm{EEA}}\,
\mathrm{diag}\!\left(
m_{\mathrm{far}}
+
\left(m_{\min}-m_{\mathrm{far}}\right)
\left[
\frac{1}{2}
\tanh\!\left(
-\alpha_m
\left(
\mathrm{dist}(\mathbf{x}^{\mathrm{eef}},\mathbf{x}^{\mathrm{eef}\star})
-
r_m
\right)
\right)
+
\frac{1}{2}
\right]
\right).
\]
For EEA, we set $m_{\mathrm{far}}=0$, so the attractor is suppressed when the target is far and smoothly activates as the target becomes reachable by the arm.
Because the EEA Jacobian is taken with respect to the full configuration $\mathbf{q}$, this term can also induce helpful base motion through the full base-to-end-effector kinematic chain.

\subsection{Base Attractor (BA)}

The BA moves the holonomic base toward a target derived from the same end-effector goal $\mathbf{T}^{\star}_{\mathrm{ee}}$.
This allows the controller to position the base automatically without requiring the policy to specify an explicit base goal.
The BA task variable is
\[
\mathbf{x}^{\mathrm{base}}
=
\left[
x_{\mathrm{base}},
y_{\mathrm{base}},
\theta_{\mathrm{base}}
\right]
\in SE(2).
\]
We derive a base target $\mathbf{x}^{\mathrm{base}\star}\in SE(2)$ from $\mathbf{T}^{\star}_{\mathrm{ee}}$: its translation is the planar projection of the end-effector target position, and its yaw is chosen so that the front of the base faces the target.

The BA uses the same attractor form as EEA, but in the $SE(2)$ base task space.
With
\[
\mathbf{e}_{\mathrm{base}}
=
\mathbf{x}^{\mathrm{base}}-\mathbf{x}^{\mathrm{base}\star},
\qquad
\rho_{\mathrm{base}}=\|\mathbf{e}_{\mathrm{base}}\|,
\]
the BA acceleration policy is
\[
\mathbf{a}_{\mathrm{BA}}
=
-k_{\mathrm{BA}}\tanh(\alpha_f \rho_{\mathrm{base}})
\frac{\mathbf{e}_{\mathrm{base}}}{\rho_{\mathrm{base}}+\epsilon}
-
d_{\mathrm{BA}}(\rho_{\mathrm{base}})\dot{\mathbf{x}}^{\mathrm{base}},
\qquad
\mathbf{f}_{\mathrm{BA}}
=
-\mathbf{M}_{\mathrm{BA}}\mathbf{a}_{\mathrm{BA}}.
\]
The BA metric uses the same smooth gate:
\[
\mathbf{M}_{\mathrm{BA}}
=
\eta_{\mathrm{BA}}\,
\mathrm{diag}\!\left(
m_{\mathrm{far}}
+
\left(m_{\min}-m_{\mathrm{far}}\right)
\left[
\frac{1}{2}
\tanh\!\left(
-\alpha_m
\left(
\mathrm{dist}(\mathbf{x}^{\mathrm{base}},\mathbf{x}^{\mathrm{base}\star})
-
r_m
\right)
\right)
+
\frac{1}{2}
\right]
\right).
\]
For BA, we set $m_{\min}=0$, so the base attractor is active when the end-effector target is far but smoothly vanishes once the target enters the arm workspace.
This prevents the base from driving directly onto the end-effector target when the arm can reach it more precisely.

\subsection{Posture Regularization (PR)}

PR is a weak joint-space attractor on the Franka arm that resolves kinematic redundancy without overriding the primary reaching terms.
It biases the arm toward a nominal configuration while using a much smaller isotropic metric than EEA and BA.

The PR task space is
\[
\mathbf{x}^{\mathrm{PR}}=\mathbf{q}_{\mathrm{Franka}},
\qquad
\mathbf{x}^{\mathrm{PR}\star}=\mathbf{q}_{\mathrm{nom}},
\qquad
\dot{\mathbf{x}}^{\mathrm{PR}}=\dot{\mathbf{q}}_{\mathrm{Franka}},
\]
with constant metric
\[
\mathbf{M}_{\mathrm{PR}}=\eta_{\mathrm{PR}}\mathbf{I}.
\]
Let $\mathbf{e}_{\mathrm{PR}}=\mathbf{x}^{\mathrm{PR}}-\mathbf{x}^{\mathrm{PR}\star}$ and $\rho_{\mathrm{PR}}=\|\mathbf{e}_{\mathrm{PR}}\|$.
We use the forcing conical attractor
\[
\mathbf{a}_{\mathrm{PR}}
=
-\|\dot{\mathbf{x}}^{\mathrm{PR}}\|^2
k_{\mathrm{PR}}
\tanh(\alpha_{\mathrm{PR}}\rho_{\mathrm{PR}})
\frac{\mathbf{e}_{\mathrm{PR}}}{\rho_{\mathrm{PR}}+\epsilon},
\qquad
\mathbf{f}_{\mathrm{PR}}
=
-\mathbf{M}_{\mathrm{PR}}\mathbf{a}_{\mathrm{PR}}.
\]
Since $\eta_{\mathrm{PR}}$ is small, PR acts as a redundancy-resolution prior while allowing reaching, collision-avoidance, and gaze terms to dominate when needed.

\subsection{Joint-Limit Repulsion (JLR)}

JLR prevents the Franka arm and ARX camera arm from approaching joint limits.
For each joint-limit boundary, we define a scalar task coordinate $s$ as the signed distance to the limit, where $s>0$ is feasible and decreasing $s$ indicates motion toward the limit.
The fabric activates when the joint is either close to the limit or moving toward it:
\[
\gamma_{\mathrm{JLR}}
=
\mathbf{1}
\left[
s < s_{\mathrm{break}}
\;\vee\;
\dot{s} < \dot{s}_{\mathrm{break}}
\right].
\]
Its metric grows rapidly near the limit,
\[
M_{\mathrm{JLR}}(s,\dot{s})
=
\gamma_{\mathrm{JLR}}
\frac{\eta_{\mathrm{JLR}}}
{
\max(s-s_{\mathrm{off}}, s_{\min})^2
},
\]
where $s_{\mathrm{off}}$ controls when the metric starts increasing and $s_{\min}$ caps the maximum metric value for numerical stability.

We use the forcing variant of the repulsion fabric, which applies acceleration away from the limit with damping when the joint is moving toward it:
\[
a_{\mathrm{JLR}}
=
k_{\mathrm{JLR}}
-
\mathbf{1}[\dot{s}\leq 0]\,
d_{\mathrm{JLR}}\dot{s},
\qquad
f_{\mathrm{JLR}}
=
-M_{\mathrm{JLR}}a_{\mathrm{JLR}}.
\]
This term has little effect away from limits, but becomes dominant near joint boundaries, preventing the WBC from using configurations with poor joint-limit margin.

\subsection{Body Repulsion (BR)}

BR provides self-collision and environment-collision avoidance.
We approximate the robot body with a set of collision spheres with centers
$\{\mathbf{x}_i^{\mathrm{BR}}\}_{i=1}^{N_s}$ and radii $\{r_i\}_{i=1}^{N_s}$.
Environment geometry is represented by local primitives, including cuboids, capsules, and spheres.
At each control step, we compute batched signed-distance queries between robot spheres and environment primitives, as well as selected robot sphere pairs for self-collision.
These distance queries are implemented with Warp kernels over environments, robot spheres, and obstacle primitives, producing closest distances and closest points in parallel.

For each robot sphere, let $d_i$ be the signed distance to its closest obstacle or self-collision partner, and let $\mathbf{p}_i$ be the closest point on that object.
We define the contact direction
\[
\mathbf{n}_i
=
\frac{\mathbf{p}_i-\mathbf{x}_i^{\mathrm{BR}}}
{\|\mathbf{p}_i-\mathbf{x}_i^{\mathrm{BR}}\|+\epsilon}.
\]
A sphere becomes active when it is within the engagement distance:
\[
\gamma_i
=
\mathbf{1}[d_i \leq d_{\mathrm{engage}}].
\]
For active spheres, BR uses a rank-one metric aligned with the collision normal,
\[
\mathbf{M}^{\mathrm{BR}}_i
=
\gamma_i
\frac{\eta_{\mathrm{BR}}}{\max(d_i,d_{\min})^2}
\mathbf{n}_i\mathbf{n}_i^\top .
\]
The repulsion direction $\mathbf{c}_i$ points away from the obstacle; when the sphere center is outside the obstacle, $\mathbf{c}_i=-\mathbf{n}_i$, and when it is inside, the direction is flipped to push the sphere outward.
The desired repulsive acceleration is
\[
\mathbf{a}^{\mathrm{BR}}_i
=
a_{\mathrm{BR}}\mathbf{c}_i
-
d_{\mathrm{BR}}\dot{\mathbf{x}}^{\mathrm{BR}}_i,
\qquad
\mathbf{f}^{\mathrm{BR}}_i
=
-\mathbf{M}^{\mathrm{BR}}_i\mathbf{a}^{\mathrm{BR}}_i .
\]
Inactive spheres have zero metric and force.
Each active sphere fabric is pulled back through the sphere-center Jacobian, so local repulsion at collision spheres induces whole-body motion of the base, arm, and neck.
This allows the same fabric formulation to handle both self-collision and scene-obstacle avoidance in a batched, environment-aware manner.

\subsection{Gaze Control Attractor (GCA)}

GCA provides active vision for the actuated camera arm.
Given a 3D look-at point $\mathbf{p}^{\star}_{\mathrm{gaze}}$, we construct a target camera pose whose position is the current camera position and whose orientation points the camera optical axis toward $\mathbf{p}^{\star}_{\mathrm{gaze}}$.
Let $\mathbf{p}_{\mathrm{cam}}(\mathbf{q})$ be the current camera position and define the desired viewing direction
\[
\mathbf{u}^{\star}_{\mathrm{gaze}}
=
\frac{
\mathbf{p}^{\star}_{\mathrm{gaze}}-\mathbf{p}_{\mathrm{cam}}(\mathbf{q})
}{
\|\mathbf{p}^{\star}_{\mathrm{gaze}}-\mathbf{p}_{\mathrm{cam}}(\mathbf{q})\|+\epsilon
}.
\]
We then construct a target orientation $\mathbf{R}^{\star}_{\mathrm{cam}}$ whose optical axis aligns with $\mathbf{u}^{\star}_{\mathrm{gaze}}$, and define
\[
\mathbf{T}^{\star}_{\mathrm{cam}}
=
(\mathbf{R}^{\star}_{\mathrm{cam}}, \mathbf{p}_{\mathrm{cam}}(\mathbf{q})).
\]
The GCA is implemented as a frame attractor analogous to the EEA, using the same keypoint pose representation and attractor force, but applied to the camera frame:
\[
\mathbf{x}^{\mathrm{cam}}
=
\kappa(\mathbf{T}_{\mathrm{cam}}(\mathbf{q})),
\qquad
\mathbf{x}^{\mathrm{cam}\star}
=
\kappa(\mathbf{T}^{\star}_{\mathrm{cam}}).
\]
This fabric rotates the actuated camera so that its optical axis tracks the look-at point, improving object visibility under partial observability.

Since the look-at target constrains only camera orientation, the ARX camera arm remains kinematically redundant, especially when combined with the mobile base degrees of freedom.
We therefore add a weak ARX joint-space posture attractor, analogous to PR, to regularize the remaining degrees of freedom:
\[
\mathbf{x}^{\mathrm{GCA\text{-}PR}}=\mathbf{q}_{\mathrm{ARX}},
\qquad
\mathbf{x}^{\mathrm{GCA\text{-}PR}\star}=\mathbf{q}_{\mathrm{ARX,nom}}.
\]
This weak regularizer stabilizes the neck posture while allowing the primary gaze fabric to dominate camera orientation.

\subsection{Fabric Energization}
After all fabric terms are composed, we apply the standard energization step from geometric fabrics~\cite{geometric_fabrics}. In our implementation, we use a joint-space Euclidean energy over the WBC configuration,
\[
E(\dot{\mathbf{q}})
=
\frac{1}{2}
\dot{\mathbf{q}}^\top
\mathbf{M}_{E}
\dot{\mathbf{q}},
\]
where $\mathbf{M}_{E}=m_E\mathbf{I}$ is a constant diagonal energy metric with isotropic mass $m_E$. 
Since $\mathbf{M}_{E}$ is constant, the corresponding Euler--Lagrange force is zero. The energy therefore only defines the kinetic energy used for energization, while the behavior itself is determined by the composed fabric acceleration. We apply energization to the raw acceleration produced by the root fabric. This stabilizes the composed WBC dynamics while preserving the geometry induced by the reaching, posture, gaze, joint-limit, and collision-avoidance fabrics \cite{geometric_fabrics}.

\section{Local RL Expert Training Details}
\label{sec:local_rl_details}
We train a suite of local RL experts for contact-rich object interaction. Since the WBC handles scene-aware reaching, each expert can be trained in a compact local scene with the end effector initialized near the object. Each expert specializes in one local interaction type, but all experts share the same action and observation structure, allowing them to be reused across many full-scene layouts during distillation.

\subsection{Environment Design}
\begin{figure}[htbp]
    \centering
    \includegraphics[width=\linewidth]{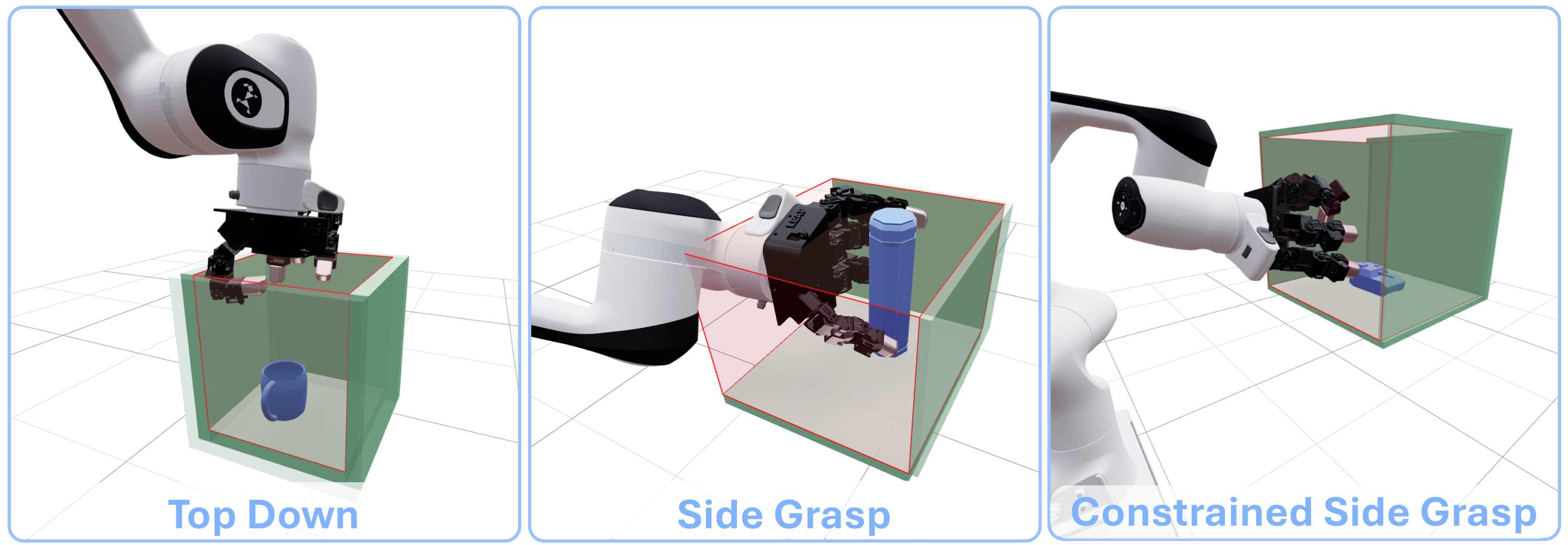}
    \caption{Visualization of the top down, side grasp, and constrained side grasp local expert policies.}
    \label{fig:local_policies_vis}
\end{figure}

Each expert environment contains a target object, a support surface, and a local interaction region that defines the nearby geometric constraints, such as an open tabletop, box, or shelf. The expert learns to approach the object within this region, establish a grasp, lift the object to a goal. See Fig.~\ref{fig:local_policies_vis} for the local expert training environments.

\begin{figure}[htbp]
    \centering
    \includegraphics[width=\linewidth]{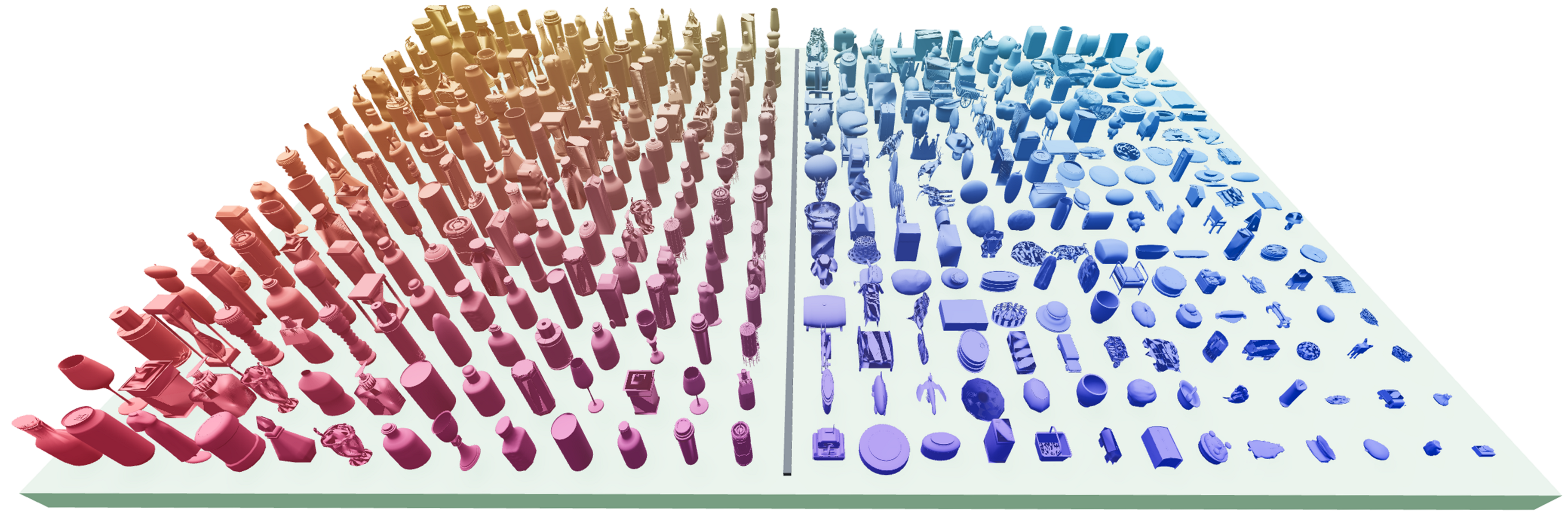}
    \caption{\textbf{Left:} Elongated objects for training the side-grasp expert and constrained side-grasp expert. \textbf{Right:} Objects for training the top-down and constrained side-grasp experts.}
    \label{fig:sim_object_set}
\end{figure}

\textbf{Objects.}
Objects are selected from Objaverse~\cite{deitke2022objaverseuniverseannotated3d}. For each object, we sample a bounding-box size and scale the mesh accordingly, adjusting its center of mass. The top-down and constrained side-grasp experts use $304$ objects with dimensions sampled from $[0.03,0.03,0.05]$ to $[0.12,0.12,0.15]$~m. The side-grasp expert uses $195$ elongated objects with dimensions sampled from $[0.03,0.03,0.15]$ to $[0.12,0.12,0.25]$~m.  See Fig.~\ref{fig:sim_object_set}.

\textbf{Resets.}
For all experts, we sample an object pose inside the local interaction region and an end-effector pose close to it, then run the WBC to convergence to obtain a collision-free, kinematically feasible robot configuration. We add initialization noise of $[-0.20,0.20]$~rad to each Franka joint and $[-0.50,0.50]$~rad to each LEAP joint, clipped to joint limits. This prevents overfitting to a single reset posture while keeping the initialization consistent with the intended grasp family. Environments reset on timeout or hard failure, where hard failure occurs if the object drops $0.1$~m below the support surface.

\paragraph{Top-down grasp expert.}
The top-down expert grasps objects in a box from above. For box scenes, we sample an open-top cuboid with inner size $[0.25,0.25,0.15]$ to $[0.50,0.50,0.25]$~m, wall thickness $0.01$~m, and center $x=0.5,y=0$. The interaction region is the box interior, with four side faces being the collision free boundary. The end effector is initialized from a range of $[0.15, 0.35]$~m above the box opening. Objects are sampled with $0.07$~m wall clearance. The end effector faces the box center with at most $0.34$~rad tilt, and the lift target is placed $0.20$~m above the box top at the box center.

\paragraph{Side-grasp expert.}
The side-grasp expert laterally grasps elongated objects. We sample two walls, with one on the lateral side of the object opposite to the end effector's side and one behind the object. Each wall has a height of $[0.15, 0.35]$~m and a width of $[0.15, 0.4]$~m, thickness of $0.01$~m. The interior region is the cubic region inside the walls, with the two sides that the walls occupy being the collision free boundary. The end effector is initialized $[0.15, 0.35]$~m from the interaction region. Objects are sampled with $0.07$~m wall clearance. The end effector faces the box center with at most $0.34$~rad tilt, and the lift target is placed $0.20$~m above the box top at the box center.

\paragraph{Constrained side-grasp expert.}
The constrained side-grasp expert handles shelf-like side extraction. We sample a shelf with inner size $[0.20,0.25,0.25]$ to $[0.35,0.50,0.40]$~m and wall thickness $0.01$~m, consisting of bottom, top, side, and back walls with the negative-$x$ face open. The interaction region is the shelf interior. Excluding the opening face and bottom face, the other four faces are the collision free boundaries. Objects are sampled with $0.07$~m wall clearance. The end effector is initialized $[0.15, 0.35]$~m outside the lateral opening of the shelf. The hand faces the object side with yaw, roll, and pitch perturbations up to $45^\circ$. The goal is placed outside the shelf, $0.10$~m beyond the opening at shelf mid-height, requiring the expert to grasp and extract the object without colliding with the shelf.

\subsection{Action Space Formulation}

All experts use a shared end-effector-frame action space, which removes unnecessary global scene information and reduces action dimensionality relative to direct joint control. At each step, the policy outputs
\[
    a = [a^p, a^R, a^h] \in [-1,1]^{22},
\]
where $a^p\in\mathbb{R}^3$ is end-effector translation, $a^R\in\mathbb{R}^3$ is end-effector rotation, and $a^h\in\mathbb{R}^{16}$ is the LEAP hand action. These are converted into deltas:
\[
    \Delta p^{\mathrm{eef}} = \alpha_p \Delta t\, a^p,\qquad
    \Delta \theta^{\mathrm{eef}} = \alpha_R \Delta t\, a^R,
\]
\[
    \Delta q^h
    =
    \left(\alpha_h \Delta t\, a^h\right)
    \odot
    \left(q_{\max}^h - q_{\min}^h\right),
\]
with $\alpha_p=0.6$, $\alpha_R=0.6$, $\alpha_h=3.0$, and $\Delta t=0.01667$. The absolute targets are
\[
    p^* = p^w + R^w_{\mathrm{eef}}\Delta p^{\mathrm{eef}},
    \qquad
    R^* = R^w_{\mathrm{eef}}\Phi(\Delta \theta^{\mathrm{eef}}),
\]
\[
    q^{h,*}
    =
    \operatorname{clip}
    \left(
    q^h+\Delta q^h,\,
    q_{\min}^h,\,
    q_{\max}^h
    \right),
\]
where $R^w_{\mathrm{eef}}$ maps vectors from the end-effector frame to the world frame and $\Phi(\cdot)$ maps axis-angle vectors to rotation matrices. The WBC tracks $(p^*,R^*)$ and outputs an arm joint target $q^{a,*}$. The simulator receives the final position target $q^*=[q^{a,*},q^{h,*}]$, tracked by the low-level position controller.

\subsection{Observation Space Formulation}

\paragraph{Actor observation.}
All experts share the same actor observation. Since experts operate in local interaction regions, all translational quantities are expressed in the current end-effector frame, including fingertip positions, object pose, goal pose, and local scene-box pose. Rotations use the continuous 6D representation. Table~\ref{tab:expert_obs_dim} lists the observation terms, with total dimension $61$.

\begin{table}[h]
\centering
\small
\setlength{\tabcolsep}{4pt}
\begin{tabular}{lc}
\toprule
\textbf{State Term} & \textbf{Dimension} \\
\midrule
LEAP hand joint positions      & 16 \\
Relative fingertip positions   & 12 \\
Scene-box bottom-center position & 3 \\
Scene-box dimensions           & 3 \\
Scene-box rotation             & 6 \\
Object position                & 3 \\
Object rotation                & 6 \\
Goal position                  & 3 \\
Goal rotation                  & 6 \\
Object bounding-box extent     & 3 \\
\midrule
\textbf{Single-step total dim} & \textbf{61} \\
\bottomrule
\end{tabular}
\vspace{1mm}
\caption{
Actor observation dimensions for local RL experts. Translational quantities are expressed in the current end-effector frame, and rotations use the continuous 6D representation.
}
\label{tab:expert_obs_dim}
\end{table}

The scene-box terms encode the local cuboid interaction region, including side walls, top walls, and shelf-like openings. This representation avoids absolute world-frame positions, making local experts reusable across different base placements during distillation. Object extent disambiguates scale and helps adapt grasp aperture and approach distance.

\paragraph{Critic state.}
We use an asymmetric critic with privileged robot and task information. In addition to the actor observation, the critic receives full robot proprioception and world-frame end-effector state, improving value estimation while keeping the actor observation compact and reusable. Table~\ref{tab:expert_critic_dim} lists the critic observation terms, with total dimension 106.

\begin{table}[h]
\centering
\small
\setlength{\tabcolsep}{4pt}
\begin{tabular}{lc}
\toprule
\textbf{State Term} & \textbf{Dimension} \\
\midrule
Full robot joint positions & 23 \\
Full robot joint velocities & 23 \\
World-frame end-effector position & 3 \\
World-frame end-effector rotation & 6 \\
End-effector linear and angular velocity & 6 \\
Relative fingertip positions & 12 \\
Scene-box bottom-center position & 3 \\
Scene-box dimensions & 3 \\
Scene-box rotation & 6 \\
Object position & 3 \\
Object rotation & 6 \\
Goal position & 3 \\
Goal rotation & 6 \\
Object bounding-box extent & 3 \\
\midrule
\textbf{Single-step total dim} & \textbf{106} \\
\bottomrule
\end{tabular}
\vspace{1mm}
\caption{
Critic-state dimensions for local RL experts. The critic receives privileged robot proprioception and world-frame end-effector state in addition to the actor observation.
}
\label{tab:expert_critic_dim}
\end{table}

\subsection{Reward Terms Formulation}

All experts use dense rewards encouraging the hand to approach the object, establish contact, lift the object, and move it to an expert-specific goal. Let $\mathbb{I}[\cdot]$ be an indicator, $p_o$ the object center or grasp point, and $\mathcal{P}_h$ the palm and fingertip keypoints. We define
\[
d_{\mathrm{hand,obj}}
=
\max_{p\in\mathcal{P}_h}\|p-p_o\|_2,
\]
which encourages the palm and all fingertips to remain close to the object. The lift condition $\mathrm{lift}$ is true when the object is no longer in contact with the table and $d_{\mathrm{hand,obj}}\leq0.15$~m. This differs from the task success criterion, which requires lifting the object by at least $0.10$~m and moving it to the goal. We use $a_{\mathrm{reg}}$ for normalized action regularization. Since the WBC generates collision-free actions, local RL rewards do not include explicit collision penalties.

For experts with orientation goals, object-goal alignment is measured using a point-matching error:
\[
d_{\mathrm{pm}}
=
\frac{1}{5}
\sum_{i=1}^{5}
\left\|
(R_c x_i + p_c) - (R_t x_i + p_t)
\right\|_2,
\]
where
\[
x_i \in
\left\{
(d,0,0),\,
(-d,0,0),\,
(0,0,0),\,
(0,d,0),\,
(0,-d,0)
\right\},
\qquad d=0.2.
\]
This captures both translation and orientation error through transformed local points.

\paragraph{Top-down grasp expert.}
The top-down expert has no goal orientation, so $d_{\mathrm{goal}}=\|p_o-p_{\mathrm{goal}}\|_2$. Its reward is
\[
R_{\mathrm{top}} =
1.0 r_{\mathrm{hand,obj}}
+ 4.0 r_{\mathrm{obj,goal}}
+ 2.0 r_{\mathrm{lift}}
+ 0.5 r_{\mathrm{curl}}
- 0.001 r_{\mathrm{act}},
\]
where
\[
r_{\mathrm{hand,obj}}=\exp(-5.0d_{\mathrm{hand,obj}}),\quad
r_{\mathrm{obj,goal}}=\mathbb{I}[\mathrm{lift}]\exp(-5.0d_{\mathrm{goal}}),
\]
\[
r_{\mathrm{lift}}=\mathbb{I}[\mathrm{lift}],\quad
r_{\mathrm{curl}}=\mathbb{I}[d_{\mathrm{hand,obj}}\leq0.15]\exp(-0.05d_{\mathrm{curl}}),\quad
r_{\mathrm{act}}=\|a_{\mathrm{reg}}\|_2^2.
\]
Here $d_{\mathrm{curl}}$ is squared deviation from a reference curled-finger posture.

\paragraph{Side-grasp expert.}
The side-grasp expert uses goal orientation $q_{\mathrm{goal}}=[-0.7071,0,0,0.7071]$ and $d_{\mathrm{goal}}=d_{\mathrm{pm}}((R_o,p_o),(R_{\mathrm{goal}},p_{\mathrm{goal}}))$. It also includes a hand-orientation error
\[
d_{\mathrm{hand,ori}}
=
\frac{1}{5}
\sum_{i=1}^{5}
\left\|
R_h x_i - R_{\mathrm{goal}}x_i
\right\|_2.
\]
The reward is
\[
R_{\mathrm{side}} =
1.0 r_{\mathrm{hand,obj}}
+ 2.0 r_{\mathrm{hand,ori}}
+ 4.0 r_{\mathrm{obj,goal}}
+ 2.0 r_{\mathrm{lift}}
+ 0.5 r_{\mathrm{curl}}
- 0.001 r_{\mathrm{act}},
\]
where $r_{\mathrm{hand,ori}}=\exp(-4.0d_{\mathrm{hand,ori}})$ and the remaining terms are defined as above.

\paragraph{Constrained side-grasp expert.}
The constrained side-grasp expert uses goal orientation $q_{\mathrm{goal}}=[0.5,-0.5,0.5,-0.5]$ and $d_{\mathrm{goal}}=d_{\mathrm{pm}}$. Its reward is
\[
R_{\mathrm{shelf}} =
1.0 r_{\mathrm{hand,obj}}
+ 4.0 r_{\mathrm{obj,goal}}
+ 2.0 r_{\mathrm{lift}}
+ 0.5 r_{\mathrm{curl}}
- 0.001 r_{\mathrm{act}}.
\]

\subsection{Optimization Algorithm}

We train each expert with asymmetric actor-critic PPO in Isaac Gym using $8192$ environments on a single NVIDIA A6000 GPU until convergence, which takes approximately $36$ hours. The actor receives the compact local observation, while the critic receives the privileged state described above. This improves value estimation without allowing the actor to rely on information unavailable during distillation. Table~\ref{tab:expert_ppo_hyperparams} lists the shared PPO hyperparameters.

\begin{table}[h]
\centering
\small
\setlength{\tabcolsep}{4pt}
\begin{tabular}{ll}
\toprule
\textbf{Parameter} & \textbf{Value} \\
\midrule
Algorithm & PPO continuous control \\
Policy model & Continuous actor with learned log standard deviation \\
Actor-critic network & Shared trunk, \texttt{separate=False} \\
MLP hidden units & $[512,256,128,64]$ \\
Activation & ELU \\
Input normalization & Enabled \\
Value normalization & Enabled \\
Advantage normalization & Enabled \\
Value bootstrap & Enabled \\
Reward scale & $1.0$ \\
Discount factor $\gamma$ & $0.99$ \\
GAE parameter $\lambda$ & $0.95$ \\
Learning rate & $5\times10^{-4}$ \\
Learning-rate schedule & Adaptive \\
KL threshold & $0.008$ \\
PPO clip range & $0.2$ \\
Horizon length & $32$ steps \\
Minibatch size & $8192$ \\
Mini-epochs per rollout & $5$ \\
Critic coefficient & $4.0$ \\
Entropy coefficient & $0.0$ \\
Gradient norm clipping & $1.0$ \\
Truncate gradients & Enabled \\
Value clipping & Enabled \\
Action clipping & Enabled \\
Action bounds loss coefficient & $10^{-4}$ \\
Sequence length & $4$ \\
Save best after & $200$ epochs \\
Checkpoint save frequency & $200$ epochs \\
\bottomrule
\end{tabular}
\vspace{1mm}
\caption{PPO hyperparameters used to train the local RL experts.}
\label{tab:expert_ppo_hyperparams}
\end{table}

The privileged central-value critic uses the same minibatch size, mini-epochs, learning rate, KL threshold, value clipping, input normalization, and gradient truncation. Its MLP also uses hidden units $[512,256,128,64]$ with ELU activations.

\subsection{Randomization}

We apply domain randomization during expert training. Unless otherwise noted, randomization samples are refreshed every $720$ simulation steps. Randomization magnitudes are linearly scheduled over $200{,}000$ simulation steps via $\min(t/200{,}000,1)$. Setup-only parameters, such as object mass, are sampled when each environment instance is created. See Table~\ref{tab:expert_domain_randomization} for a detailed list of parameters.

\begin{table}[h]
\centering
\small
\setlength{\tabcolsep}{4pt}
\begin{tabular}{lll}
\toprule
\textbf{Parameter} & \textbf{Initial Setting} & \textbf{Terminal Setting} \\
\midrule
Observation white noise, top/shelf & $\mathcal{N}(0,0)$ & $\mathcal{N}(0,0.02)$ \\
Observation white noise, side-table & $\mathcal{N}(0,0)$ & $\mathcal{N}(0,0.002)$ \\
Observation correlated noise & $\mathcal{N}(0,0)$ & $\mathcal{N}(0,0.001)$ \\
Action white noise, top/shelf & $\mathcal{N}(0,0)$ & $\mathcal{N}(0,0.10)$ \\
Action white noise, side-table & $\mathcal{N}(0,0)$ & $\mathcal{N}(0,0.05)$ \\
Action correlated noise & $\mathcal{N}(0,0)$ & $\mathcal{N}(0,0.015)$ \\
Gravity perturbation & $\mathcal{N}(0,0)$ & $\mathcal{N}(0,0.4)$ \\
Robot joint damping scaling & $\mathrm{Log}\mathcal{U}(1,1)$ & $\mathrm{Log}\mathcal{U}(0.3,3.0)$ \\
Robot joint stiffness scaling & $\mathrm{Log}\mathcal{U}(1,1)$ & $\mathrm{Log}\mathcal{U}(0.75,1.5)$ \\
Robot joint lower-limit perturbation & $\mathcal{N}(0,0)$ & $\mathcal{N}(0,0.01)$ \\
Robot joint upper-limit perturbation & $\mathcal{N}(0,0)$ & $\mathcal{N}(0,0.01)$ \\
Robot friction scaling & $\mathcal{U}(1,1)$ & $\mathcal{U}(0.7,1.3)$ \\
Object mass scaling & $\mathcal{U}(1,1)$ & $\mathcal{U}(0.2,0.8)$ \\
Object friction scaling & $\mathcal{U}(1,1)$ & $\mathcal{U}(0.7,1.3)$ \\
Table friction scaling & $\mathcal{U}(1,1)$ & $\mathcal{U}(0.7,1.3)$ \\
\bottomrule
\end{tabular}
\vspace{1mm}
\caption{
Domain randomization parameters for local expert training. Gaussian entries report zero-mean perturbations with the listed terminal standard deviation. Friction randomization uses $250$ buckets for the robot, object, and table.
}
\label{tab:expert_domain_randomization}
\end{table}

We additionally apply random object wrenches when the end effector is within $0.15$~m of the object or after the object is lifted. A new wrench is sampled every $1.0$~s. The linear force direction is sampled uniformly, with magnitude up to the current curriculum maximum; torque direction is random, with magnitude equal to force magnitude times a torsional radius of $0.01$~m. See Table~\ref{tab:expert_wrench_randomization} for a detailed list of parameters.

\begin{table}[h]
\centering
\small
\setlength{\tabcolsep}{4pt}
\begin{tabular}{lll}
\toprule
\textbf{Parameter} & \textbf{Initial Setting} & \textbf{Terminal Setting} \\
\midrule
Object wrench force & $\mathcal{U}(0,0.5)\,\mathrm{N}$ & $\mathcal{U}(0,5.0)\,\mathrm{N}$ \\
Wrench resampling interval & $1.0\,\mathrm{s}$ & $1.0\,\mathrm{s}$ \\
Torsional radius & $0.01\,\mathrm{m}$ & $0.01\,\mathrm{m}$ \\
\bottomrule
\end{tabular}
\vspace{1mm}
\caption{Random wrench disturbance used during expert training.}
\label{tab:expert_wrench_randomization}
\end{table}

Finally, we apply staggered environment reset by randomly initializing the progress buffer at the start of training.

\section{Full-Scene Multi-Teacher Distillation Details}
\label{sec:distillation_details}

This section provides additional details for full-scene multi-teacher distillation. We distill composed teachers consisting of the vectorized WBC and local RL experts across thousands of diverse scenes. The student learns from both the WBC's scene-aware reaching behavior and the local experts' contact-rich manipulation behavior, producing a single onboard policy that adapts to object and scene geometry from partial point-cloud observations.

\subsection{Full-Scene Environment Design}

\paragraph{Scene geometry.}
Full-scene environments embed the local interaction regions used by the experts into diverse mobile-manipulation scenes. Each training environment contains a target object, support surface, local interaction region, and surrounding obstacles. The scene family is selected to match a local expert: top-down grasping in boxes, drawers, cluttered tabletops, or open containers; side grasping on cluttered tabletops; and constrained side grasping in shelf-like scenes.

For tabletop scenes, the table height is sampled from $[0.1,0.8]$~m and dimensions from $[0.5,0.6,0.05]$ to $[0.7,1.2,0.05]$~m. We add background distractors beside, behind, and below the table to increase visual diversity; these appear in the point cloud but are not simulated as rigid bodies. We also add free-space obstacles around the local interaction region as rigid bodies included in the WBC collision world. These obstacles force the student to adapt base approach and orientation around scene geometry. Additionally, obstacle cuboids are sampled near the approach path with dimensions from $[0.15,0.15,0.3]$ to $[0.4,0.4,1.7]$~m.

\begin{figure}[t]
    \centering
    \includegraphics[width=\linewidth]{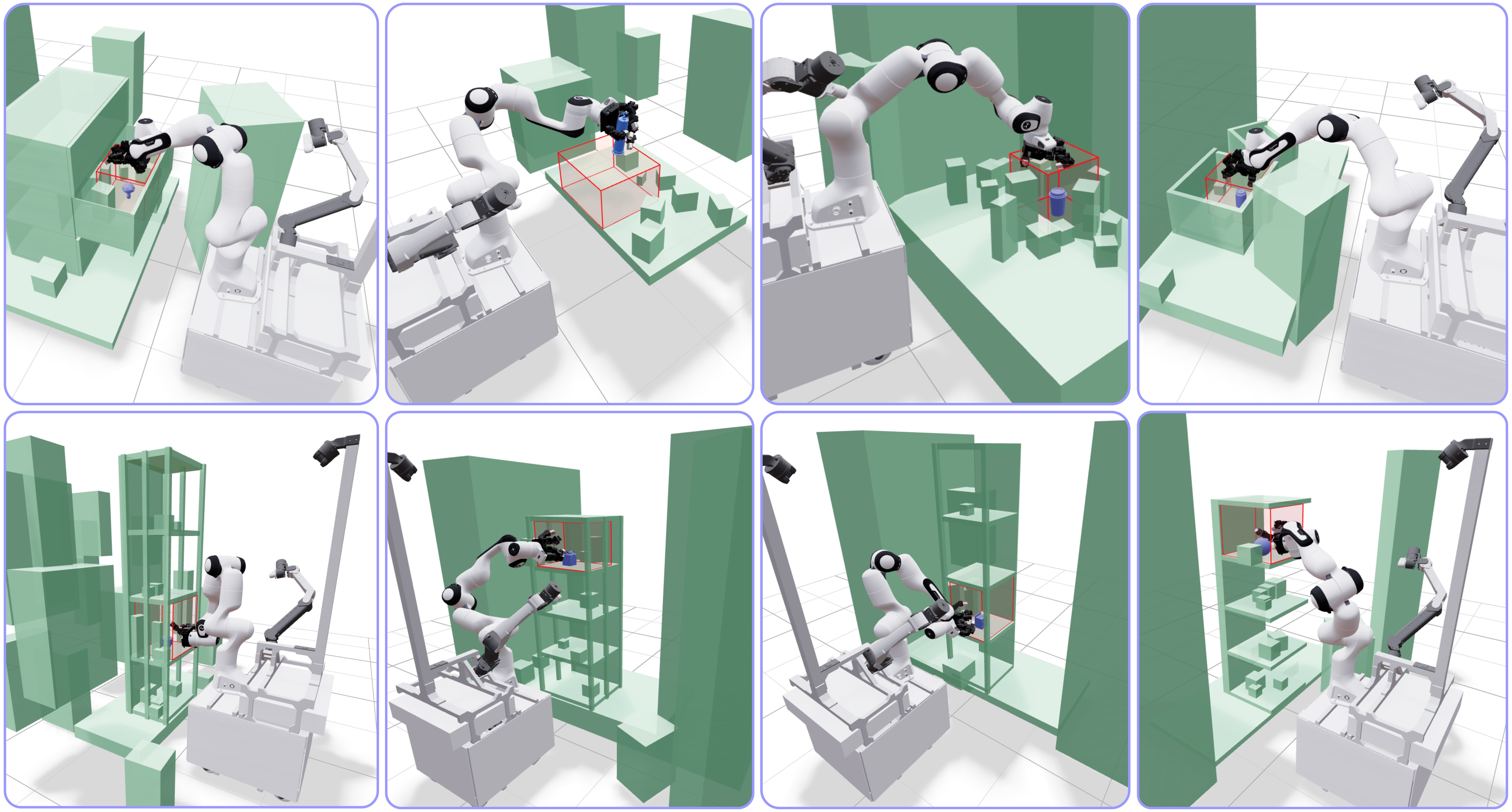}
    \caption{
\textbf{Visualization of full-scene simulation environments.}
Procedural generation produces diverse scene types, including drawers, cluttered tabletops, boxes, and shelves.
Each scene is paired with a local RL expert based on both the scene geometry and object geometry: elongated tabletop objects use the side-grasp expert, more compact tabletop objects use the top-down expert, and shelf scenes use the constrained side-grasp expert.
    }
    \label{fig:full_scene_vis}
\end{figure}

\subsection{Teacher Policies}
Each teacher rollout has two phases. During the free-space reaching phase, the WBC moves the end effector to a switching target near the object while avoiding obstacles and maintaining gaze on the target. Once the end effector reaches the switching target, the corresponding local RL expert is activated and performs contact-rich manipulation through the WBC.

The switching target depends on the expert family. For top-down scenes, it is placed $0.2$~m above the object. For side-grasp scenes, it is placed $0.2$~m laterally from the object along the local side approach direction, with height set by the object half-extent. For constrained side-grasp scenes, the target is projected to the shelf opening, shifted outward and upward by $0.15$~m, and assigned the local expert's desired approach orientation composed with the shelf frame. The student distills the resulting composed WBC--expert behavior end-to-end into a single full-body policy.

\subsection{Student Action Space}
\label{sec:student_policy_details}
The student outputs a unified full-body action
\[
a=[a^b,a^f,a^h,a^v]\in[-1,1]^{32},
\]
where $a^b\in\mathbb{R}^3$ controls the mobile base, $a^f\in\mathbb{R}^7$ the Franka arm, $a^h\in\mathbb{R}^{16}$ the LEAP hand, and $a^v\in\mathbb{R}^6$ the ARX camera neck.

\paragraph{Base command.}
The base action is interpreted as a base-frame $SE(2)$ velocity. Given current base pose $q_t^b=(x_t,y_t,\psi_t)$ and control step $\Delta t=0.01667$~s,
\[
\Delta q^b_{\mathrm{world}}
=
R_{\mathrm{SE(2)}}(\psi_t)(\Delta t\,a^b),
\qquad
q^{b,*}_{t+1}=q_t^b+\Delta q^b_{\mathrm{world}}.
\]

\paragraph{Joint commands.}
For robot part $r\in\{f,h,v\}$, the decoded joint target is
\[
q^{r,*}_{t+1}
=
q_t^r
+
\alpha_r\Delta t
(q_{\max}^r-q_{\min}^r)\odot a^r,
\qquad
\alpha_f=0.6,\ \alpha_h=3.0,\ \alpha_v=0.6.
\]
The resulting target is clipped to joint limits and sent to the simulator position controller.

\paragraph{Teacher-action encoding.}
During supervision, the composed teacher first produces a full-body target $q^{*,T}_{t+1}$ from the WBC and local expert. We convert this target back into the student's normalized action space:
\[
a^{b,T}
=
\frac{
R_{\mathrm{SE(2)}}(-\psi_t)(q^{b,*,T}_{t+1}-q_t^b)
}{\Delta t},
\]
\[
a^{r,T}
=
\frac{
q^{r,*,T}_{t+1}-q_t^r
}{
\alpha_r\Delta t(q_{\max}^r-q_{\min}^r)
},
\qquad
r\in\{f,h,v\}.
\]
The student is trained to regress this converted 32-D action.

\subsection{Student Observation Space}

The student receives partial point-cloud observations from the actuated depth camera and base-mounted LiDAR, robot proprioception, LEAP hand controller state, and a 3D target-object position estimate. Table~\ref{tab:student_obs_dim} summarizes the observation terms.

\paragraph{Target-object position estimate.}
For whole-body dexterous grasping, the student is conditioned on a 3D target-object position estimate, which specifies the desired object in cluttered scenes and serves as the center for object-centric visual crops. During training, the estimate is represented in the current robot base frame and updated autoregressively from the previous policy prediction; on object reset, it is refreshed from the true object state.

At deployment, the estimate can be initialized manually or from SAM3~\cite{carion2025sam3segmentconcepts}. For SAM3 initialization, we select the highest-probability mask, lift it to 3D, and use the point centroid as the initial target-object position. The policy then updates this estimate through an auxiliary target-position prediction. SAM3 can optionally be rerun at $3$~Hz to recover from tracking failures.

\begin{figure}[t]
    \centering
    \includegraphics[width=\linewidth]{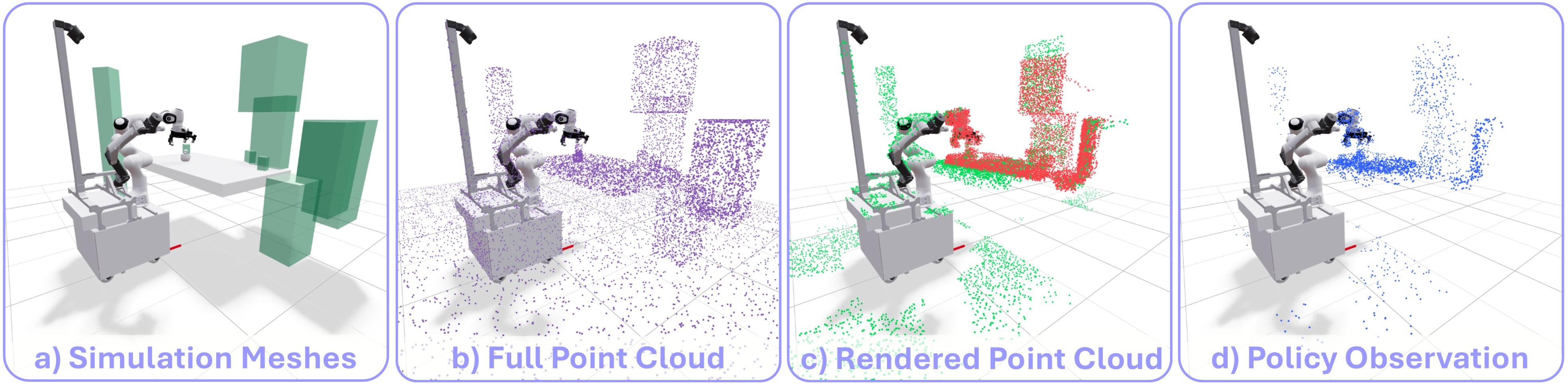}
    \caption{
    \textbf{Pipeline for generating point cloud observations for the student policy.}
    a) Each scene starts with primitive meshes in simulation.
    b) We sample points on the surface of the meshes to generate a ground-truth full point cloud.
    c) Given the full point cloud, we then use the vectorized depth camera and LiDAR renderer to produce a simulated point cloud for each sensor. The green point cloud is the simulated LiDAR points. The red point cloud is the simulated depth camera points.
    d) We finally downsample and concatenate both the simulated LiDAR and depth camera point clouds into a final point cloud that we feed as observation to the student policy.
    }
    \label{fig:pcd_viz}
\end{figure}

\paragraph{Point-cloud construction.}
We simulate partial onboard observations by rendering a global point cloud into sensor-like depth and LiDAR clouds. Each environment constructs
\[
P_{\mathrm{gt}}=[P_{\mathrm{scene}},P_{\mathrm{robot}}],
\]
then renders viewpoint-dependent observations
\[
P_{\mathrm{depth}}=\mathrm{RenderDepth}(P_{\mathrm{gt}},T_{\mathrm{cam}}),
\qquad
P_{\mathrm{lidar}}=\mathrm{RenderLidar}(P_{\mathrm{gt}},T_{\mathrm{lidar}}),
\]
Rendering is parallelized across environments. Details on the vectorized depth-camera and LiDAR renderers are provided in Appendix~\ref{sec:vectorized_sensor_rendering}. Simulated LiDAR and depth camera point clouds are visualized in Fig.~\ref{fig:pcd_viz}.

\paragraph{Local crops.}
The policy receives both scene-level and object-level geometry. From each rendered depth and LiDAR cloud, we extract: (1) a cylindrical base crop centered at the Franka base with radius $1.5$~m, downsampled to $N_{\mathrm{base}}=1024$ points, and (2) a spherical object crop centered at the current target-object estimate with radius $0.5$~m, downsampled to $N_{\mathrm{obj}}=512$ points. The base crop provides broader scene context for base motion and collision avoidance, while the object crop preserves denser local geometry for grasping.

The final observed cloud is
\[
P_{\mathrm{obs}}
=
[
P_{\mathrm{base}}^{\mathrm{depth}},
P_{\mathrm{base}}^{\mathrm{lidar}},
P_{\mathrm{obj}}^{\mathrm{depth}},
P_{\mathrm{obj}}^{\mathrm{lidar}}
],
\]
for a total of $3072$ points. Here $P_{\mathrm{base}}^{(\cdot)}$ denotes the base-centered crop and $P_{\mathrm{obj}}^{(\cdot)}$ denotes the object-centered crop from each sensor. We transform $P_{\mathrm{obs}}$ into the Franka-base frame before passing it to the student, removing global translation and yaw variation while preserving local scene geometry. In Fig.~\ref{fig:pcd_viz}, we visualize the scene mesh, the full scene point cloud, the rendered point clouds from the LiDAR and depth cameras, and the final cropped and down-sampled point cloud that we feed into the policy. 

\paragraph{Proprioception and hand-controller state.}
In addition to point clouds, the student receives normalized Franka, ARX, and LEAP joint positions. We also provide the LEAP hand control-target delta
\[
\Delta q_{\mathrm{hand}}
=
q^{\mathrm{target}}_{\mathrm{hand}}-q_{\mathrm{hand}}.
\]
During contact-rich manipulation, commanded hand targets can differ from realized joint positions; this residual provides useful contact information for reactive adjustment.

\begin{table}[h]
\centering
\small
\setlength{\tabcolsep}{4pt}
\begin{tabular}{lc}
\toprule
\textbf{Observation Term} & \textbf{Dimension} \\
\midrule
Observed point cloud in Franka-base frame & $3072 \times 3$ \\
Franka manipulation-arm joint positions & 7 \\
ARX vision-arm joint positions & 6 \\
LEAP hand joint positions & 16 \\
LEAP hand control-target delta & 16 \\
Target-object position estimate & 3 \\
\bottomrule
\end{tabular}
\vspace{1mm}
\caption{
Student observation terms. Joint states are normalized by joint limits. The point cloud is encoded as tokens by the PointNet encoder rather than flattened into a vector.
}
\label{tab:student_obs_dim}
\end{table}

\subsection{Student Policy Model Architecture and Distillation Objective}
\label{sec:student_architecture_and_objective}
The student is a point-cloud conditioned transformer that maps $P_{\mathrm{obs}}$, normalized robot-state tokens, and the target-object position estimate to full-body mobile-manipulation actions. A single network is shared across all teacher families during multi-teacher DAgger training.

\paragraph{Tokenization.}
The $3072$-point input cloud is encoded with a PointNet set-abstraction encoder using radius $0.1$, $64$ neighbors, and $128$ output tokens. The PointNet MLP maps XYZ features through $[3,64,64,64]$, and a projection MLP maps each output token to $124$ features. Each low-dimensional state input is encoded by a separate MLP with hidden dimensions $[128,256]$ and output dimension $124$. We use five state tokens: Franka joint state, ARX joint state, LEAP joint state, LEAP control-target delta, and target-object position estimate. A learned $4$-D type embedding is concatenated to every token, giving transformer width
\[
d_{\mathrm{model}}=124+4=128.
\]
The encoder receives $128$ point-cloud tokens and $5$ state tokens.

\paragraph{Transformer and prediction heads.}
The backbone is an encoder-decoder transformer with $6$ encoder layers, $6$ decoder layers, $8$ attention heads, and feed-forward width $1024$. The decoder uses $C$ learned action queries and one auxiliary query. Each action query predicts one 32-D full-body action, giving
\[
\hat{A}\in\mathbb{R}^{B\times C\times32}.
\]
The auxiliary query predicts a 3D update to the target-object estimate. It is masked to cross-attend only to point-cloud tokens and the previous target-object estimate token, and is prevented from attending to action queries. This encourages the auxiliary head to estimate object position from geometry rather than action outputs.

\paragraph{Target-object estimate prediction.}
The auxiliary head predicts a bounded delta
\[
\hat{\delta}_{\mathrm{obj}}\in[-1,1]^3,
\qquad
\hat{o}_{t+1}
=
o_t^{\mathrm{anchor}}
+
s_{\mathrm{obj}}\hat{\delta}_{\mathrm{obj}},
\qquad
s_{\mathrm{obj}}=0.01~\mathrm{m}.
\]
The predicted estimate is fed back as the next observation input and used as the center for the object crop. During training, we perturb the target-object estimate to prevent the policy from over-relying on perfect object anchors.

\paragraph{Distillation loss.}
For a batch, let $A^T\in\mathbb{R}^{B\times C\times32}$ be the converted teacher action chunk and $y_{\mathrm{obj}}\in\mathbb{R}^{B\times1\times3}$ the target-object position label. We optimize
\[
\mathcal{L}_{\mathrm{act}}
=
\frac{1}{32BC}
\sum_{b=1}^{B}\sum_{c=1}^{C}
\|\hat{A}_{b,c}-A^T_{b,c}\|_2^2,
\]
\[
\mathcal{L}_{\mathrm{obj}}
=
\frac{1}{3B}
\sum_{b=1}^{B}
\|\hat{y}_{\mathrm{obj},b}-y_{\mathrm{obj},b}\|_2^2,
\qquad
\mathcal{L}
=
\mathcal{L}_{\mathrm{act}}+0.1\mathcal{L}_{\mathrm{obj}}.
\]
We train with AdamW, gradient clipping at norm $1.0$, cosine learning-rate decay with warmup, and BF16 autocast on CUDA.

\paragraph{Multi-teacher DAgger.}
Each GPU rank runs an expert-specific full-scene environment and teacher while sharing the same student weights. Gradients are averaged across ranks. We train on 8 NVIDIA A6000 GPUs for approximately 60 hours, with 512 environments per rank. During rollouts, the student executes actions while the current teacher provides supervised labels, following online DAgger.

\paragraph{Teacher forcing and early reset.}
To stabilize early training, all environments execute teacher actions for the first $10$k gradient steps. The teacher-forced fraction then decays linearly to zero over the next $10$k steps. After this warmup, environments execute student actions while still collecting teacher labels. Environments reset on local failures or after sustained success: if the success condition holds for $100$ consecutive simulation steps, the episode resets early.

\subsection{Object Localization Pipeline}
\label{sec:sam3_localization}
The student policy is conditioned on an auxiliary 3D target-object position estimate, which identifies the object to be grasped and tracks the target object in cluttered scenes. At each inference step, an auxiliary policy head predicts an updated target position (Sec.~\ref{sec:student_architecture_and_objective}); this prediction is used as input at the subsequent step. During training, rollouts use the policy's predicted target position, while the auxiliary head is supervised using the ground-truth object position, which is directly available in simulation. Consequently, given an initial target-position estimate at the start of a rollout, the policy can update this estimate autoregressively throughout execution.

For real-world experiments, SAM3 provides the initial target-position estimate. For each trial, we provide a prompt that yields a valid initial detection, lift the resulting mask into 3D, and use its centroid as the initial estimate. The policy then updates the estimate at 60\,Hz, while SAM3 runs at 3\,Hz and overwrites the estimate whenever it returns a valid detection, providing additional robustness to tracking failures. If SAM3 fails during a rollout, autoregressive tracking continues. Any failures after rollout initialization are included in the reported success rate.

\subsection{Randomization}

Student distillation uses the same domain randomization as expert training, including random wrenches and staggered environment reset, but with all randomization magnitudes set to their maximum values from the start.

We additionally use object teleportation as a robustness augmentation. During an episode, the target object pose is occasionally resampled without resetting the robot, forcing the student to recover from off-distribution perception and control states while still receiving supervision from the appropriate local expert. Teleportation is applied only to a cached subset of environments with proportion $\rho_{\mathrm{tele}}=0.2$, periodically resampled.

At simulation step $t$, the teleport probability follows a cyclic curriculum. Let $i=t\bmod n_2$ and $c(t)=\min(t/N_{\mathrm{curri}},1)$. The raw probability is
\[
p_{\mathrm{raw}}(i)=
\begin{cases}
0, & i<n_0,\\
\frac{0.5}{n_1-n_0}, & n_0\leq i<n_1,\\
a(i+1-n_1)^2+\frac{0.5}{n_1-n_0}, & n_1\leq i<n_2,
\end{cases}
\]
where $a$ is chosen so the quadratic branch reaches probability $1.0$ at the end of the cycle. We use
\[
\rho_{\mathrm{tele}}=0.2,\quad
N_{\mathrm{curri}}=200{,}000,\quad
n_0=10,\quad
n_1=150,\quad
n_2=300,
\]
and set $p_{\mathrm{tele}}(t)=c(t)p_{\mathrm{raw}}(t\bmod n_2)$. If a teleport event is sampled, only the object state is reset for active cached teleport environments.

\section{Vectorized Depth Camera and LiDAR Simulation}
\label{sec:vectorized_sensor_rendering}

We implement GPU-vectorized sensor renderers that convert batched full-scene point clouds into partial onboard observations. 
Unlike conventional simulation pipelines that render directly from ground-truth meshes, we first sample complete point clouds from scene and robot geometry, then render sensor-like observations from these point clouds. 
This trades rendering fidelity for speed and GPU-memory usage, which is important for large-scale distillation. 
Across $1024$ Isaac Gym environments on a single NVIDIA A6000, rendering depth observations at $212{\times}120$ resolution and LiDAR observations with $32768$ points takes $0.07$~s and $0.25$~s, respectively. Visualization of the simulated rendered point clouds are provided in Fig.~\ref{fig:pcd_viz}.

\subsection{Vectorized Depth Camera Renderer}

The depth renderer takes a batched full-scene point cloud $P_{\mathrm{gt}}\in\mathbb{R}^{B\times N\times3}$ and camera poses $T_{\mathrm{cam}}\in\mathbb{R}^{B\times7}$, where each pose contains position and quaternion orientation. 
For each environment, we transform points into the camera frame, project them into image coordinates using field-of-view-derived intrinsics, perform z-buffering to keep only the closest point per pixel, and backproject the resulting depth image into a rendered world-frame point cloud.

We assume the camera optical axis is the $+X$ axis of the camera frame. 
Given the camera pose, we construct the world-frame camera basis $(\hat{\mathbf{u}},\hat{\mathbf{w}},\hat{\mathbf{v}})$, corresponding to horizontal, vertical, and forward/depth directions. 
For each point $\mathbf{p}$, we compute camera coordinates by projection onto this basis:
\[
\mathbf{r} = \mathbf{p} - \mathbf{p}_{\mathrm{cam}},
\qquad
x = \mathbf{r}^{\top}\hat{\mathbf{u}},
\qquad
y = \mathbf{r}^{\top}\hat{\mathbf{w}},
\qquad
z = \mathbf{r}^{\top}\hat{\mathbf{v}}.
\]
Points are then projected to pixels using intrinsics computed from the image resolution and camera field of view:
\[
u = f_x \frac{x}{z} + c_x,
\qquad
v = f_y \frac{y}{z} + c_y.
\]
We discard points outside the image bounds, behind the camera, or outside the near/far clipping range.

To approximate visibility, we use a batched z-buffer. 
For each pixel, only the point with minimum depth $z$ is retained, while farther points projected to the same pixel are removed as occluded. 
This produces a depth image $D\in\mathbb{R}^{B\times H\times W}$ with invalid pixels set to infinity. 
Since we render from point-sampled geometry rather than continuous meshes, sparse occluders can leave holes in the z-buffer and allow occluded background points to leak through. We therefore inflate occluders with a local min-depth operation over neighboring pixels, producing a more conservative visibility mask near depth boundaries.

Finally, we backproject each valid depth pixel into world coordinates:
\[
x = \frac{u-c_x}{f_x}D(u,v),
\qquad
y = \frac{v-c_y}{f_y}D(u,v),
\]
\[
\mathbf{p}_{\mathrm{world}}
=
\mathbf{p}_{\mathrm{cam}}
+
x\hat{\mathbf{u}}
+
y\hat{\mathbf{w}}
+
D(u,v)\hat{\mathbf{v}}.
\]
We optionally add small metric noise to valid points to mimic sensor noise. 
The rendered point cloud is shuffled, invalid points are sorted to the end, and the first fixed number of valid-or-padded points is returned. 
All operations are implemented as batched tensor operations over the environment dimension, allowing depth observations to be rendered in parallel across thousands of scenes.

\subsection{Vectorized LiDAR Renderer}

The LiDAR renderer also operates directly on the batched full-scene point cloud
$P_{\mathrm{gt}}\in\mathbb{R}^{B\times N\times 3}$.
Given LiDAR poses $T_{\mathrm{lidar}}\in\mathbb{R}^{B\times 7}$, we transform all points into the LiDAR frame and simulate a hemispherical scan over the LiDAR $+Z$ direction.
For each point, we compute its range and angular coordinates:
\[
r=\|\mathbf{p}_{\mathrm{lidar}}\|_2,\qquad
\phi=\mathrm{atan2}(y,x),\qquad
\theta=\arccos(z/r),
\]
where $\phi$ is the azimuth and $\theta\in[0,\pi/2]$ is the polar angle over the scanned hemisphere.
Points outside the hemisphere or outside the near/far range are discarded.

We discretize the scan into $N_{\mathrm{polar}}\times N_{\mathrm{azimuth}}$ angular bins.
Each valid point is assigned to a bin based on $(\theta,\phi)$, and we keep only the closest range in each bin, producing a first-return range image:
\[
R[b_{\theta},b_{\phi}]
=
\min_{\mathbf{p}\in \mathcal{B}(b_{\theta},b_{\phi})}
\|\mathbf{p}_{\mathrm{lidar}}\|_2 .
\]
This removes points occluded by closer geometry along the same angular ray.

Because the scene is represented by sampled points rather than continuous surfaces, sparse foreground geometry can leave holes in the angular range image, allowing farther background points to leak through near object silhouettes.
To suppress these artifacts without artificially filling empty rays, we compare each valid return to the minimum range in a local angular neighborhood.
A return is removed if it is significantly farther than a nearby closer return:
\[
R(b) >
R_{\mathrm{nbr}}(b)
+
\epsilon_{\mathrm{abs}}
+
\epsilon_{\mathrm{rel}} R_{\mathrm{nbr}}(b),
\]
where $R_{\mathrm{nbr}}(b)$ is the local neighborhood minimum.
This suppresses likely leak-through returns while preserving empty bins.

Finally, each valid angular bin is backprojected using its bin-center ray direction.
Let $(\theta_b,\phi_b)$ be the angular center of bin $b$ and
\[
\mathbf{d}_b
=
[\sin\theta_b\cos\phi_b,\,
  \sin\theta_b\sin\phi_b,\,
  \cos\theta_b]^\top
\]
be the corresponding unit ray in the LiDAR frame.
The world-frame return is
\[
\mathbf{p}^{\mathrm{world}}_b
=
\mathbf{p}_{\mathrm{lidar}}
+
\mathbf{R}_{\mathrm{lidar}} R(b)\mathbf{d}_b .
\]
Invalid bins are represented as NaNs, returns are shuffled, NaNs are sorted to the end, and the first fixed number of points is returned.
This produces a fixed-size LiDAR point cloud for each environment while preserving viewpoint-dependent visibility and occlusion.
All steps are implemented as batched tensor operations across environments.

\section{Additional Experiments and Details}

\subsection{Policy Observation Ablation Details}
\label{sec:obs_ablation_details}

We ablate the onboard observation design by retraining the student policy under three variants while keeping the rest of the distillation pipeline unchanged. We test both in simulation and in the real world.

\textbf{No active vision.}
In this baseline, the teacher policy does not use the camera gaze attractor term in the fabric WBC. Instead, the ARX camera arm is fixed at its default configuration,
\[
\mathbf{q}_{\mathrm{ARX}}
=
[0.0,\;0.785,\;0.785,\;0.0,\;0.0,\;0.0].
\]
The student therefore receives depth observations from a fixed camera pose rather than an actively controlled viewpoint. This tests whether actuating the camera to track the object is necessary for maintaining visibility during approach and interaction.

\textbf{No LiDAR input.}
In this baseline, active vision remains enabled, but the student policy does not receive LiDAR point clouds. The policy must rely only on the actuated-neck depth camera and proprioception for scene understanding. This tests whether the base-mounted LiDAR provides complementary global context for base motion and collision avoidance, especially when nearby structures fall outside the depth camera field of view or minimum sensing range.

\textbf{No simulated rendering.}
In this baseline, the policy is trained with ground-truth full point clouds of the scene and object, without passing them through the vectorized depth-camera and LiDAR rendering pipeline. Thus, the policy observes complete scene geometry during training rather than viewpoint-dependent partial observations. This tests whether training on rendered partial sensor observations is necessary for sim-to-real transfer, since the real robot only receives incomplete onboard point clouds.

\subsection{WBC Ablation Details}
\label{sec:wbc_ablation_details}

We evaluate WBC design by replacing the controller used by teacher policies during both the reaching and interaction phases. The evaluation uses the same held-out setup as Sec.~\ref{sec:sim_evals}: $8{,}000$ simulation environments and $150$ held-out objects.

\textbf{Weighted IK.}
The first baseline is the weighted differential IK controller described in Sec.~\ref{sec:weighted_ik}. It penalizes base motion more strongly than arm motion, encouraging the arm to reach whenever possible.

\textbf{Fabrics WBC without collision avoidance.}
The second baseline uses our fabric WBC without body repulsion terms. It still includes the whole-body reaching terms, including the end-effector attractor (EEA), base attractor (BA), and posture regularization, but does not include self- or environment-collision avoidance.

\textbf{Full fabrics WBC.}
The final variant is the full WBC used by COWBOY, including whole-body reaching, posture regularization, joint-limit repulsion, active gaze control, and body repulsion fabrics for self- and environment-collision avoidance.

We report grasp success and collision rate using the same definitions as Sec.~\ref{sec:sim_evals}. In addition, we report arm manipulability and joint-limit margin to quantify the quality of the whole-body motion.

For arm manipulability, we compute Yoshikawa manipulability \cite{yoshikawa1985manipulability} using the Franka arm-only end-effector Jacobian:
\[
m_t
=
\sqrt{
\det
\left(
\mathbf{J}_{\mathrm{arm}}(\mathbf{q}_t)
\mathbf{J}_{\mathrm{arm}}(\mathbf{q}_t)^\top
\right)
}.
\]
We report the average over rollout timesteps and environments:
\[
\bar{m}
=
\frac{1}{NT}
\sum_{n=1}^{N}
\sum_{t=1}^{T}
m_{n,t}.
\]

For joint-limit margin, we compute the distance from each Franka joint to its closest joint limit,
\[
\Delta q_{i,t}
=
\min
\left(
q_{i,t}-q_i^{\min},
q_i^{\max}-q_{i,t}
\right),
\]
and average across arm joints, timesteps, and environments:
\[
\overline{\Delta q}
=
\frac{1}{NTd}
\sum_{n=1}^{N}
\sum_{t=1}^{T}
\sum_{i=1}^{d}
\Delta q_{n,t,i}.
\]
A higher margin indicates that the controller keeps the arm farther from joint limits, while higher manipulability indicates that the arm remains in configurations better suited for local manipulation.